\documentclass{article}

\PassOptionsToPackage{numbers, compress}{natbib}
\usepackage[main, final]{neurips_2026}
\usepackage{graphicx}
\usepackage{subcaption}
\usepackage{booktabs} 
\usepackage[hidelinks]{hyperref}
\usepackage{makecell}

\usepackage{amsmath}
\usepackage{amssymb}
\usepackage{mathtools}
\usepackage{amsthm}
\usepackage{float}
\usepackage[capitalize,noabbrev]{cleveref}
\usepackage{tocloft}
\theoremstyle{plain}

\theoremstyle{definition}

\theoremstyle{remark}

\usepackage{amsmath,amssymb}
\usepackage{algorithm}
\usepackage{algpseudocode} 
\usepackage{multirow}

\usepackage{colortbl}

\usepackage{multirow}
\usepackage[table]{xcolor}
\definecolor{advgray}{gray}{0.9}

\definecolor{cleanrow}{RGB}{245,250,255}
\definecolor{advrow}{RGB}{255,245,245}
\definecolor{headrow}{RGB}{235,235,235}

\definecolor{cWhisperM}{RGB}{235,235,235}
\definecolor{cWav2vec}{RGB}{252,252,252}
\definecolor{cMMS}{RGB}{235,235,235}
\definecolor{cWhisperL}{RGB}{252,252,252}

\definecolor{CG}{HTML}{555555}  

\usepackage{array}
\newcolumntype{B}{>{\bfseries}l}
\newcolumntype{C}{>{\bfseries}c}

\usepackage[textsize=tiny]{todonotes}

\usepackage[utf8]{inputenc} 
\usepackage[T1]{fontenc}    
\usepackage{url}            
\usepackage{amsfonts}       
\usepackage{nicefrac}       
\usepackage{microtype}      
\usepackage{xcolor}         

\usepackage{wrapfig}

\title{TransferBreaker: Breaking Adversarial Transferability in Fine-Tuned Speech Recognition}

\author{
  Mojtaba Nafez{\large\thanks{Equal contribution.}} \\
  Idiap Research Institute \\
  \makebox[145pt][c]{École Polytechnique Fédérale de Lausanne (EPFL)} \\
  \texttt{mojtaba.nafez@epfl.ch} \\
  \And
  Aref Mousavi{\large\footnotemark[1]} \\
  Department of Computer Engineering \\
  Sharif University of Technology \\
  \texttt{aref.mousavi.ac@gmail.com} \\
  \AND
  Mohammad Ebrahim Mahdavi \\
  Department of Computer Engineering \\
  University of Isfahan \\
  \texttt{mhm.ebrh.mahdavi@gmail.com} \\
  \And
  Mobina Poulaei \\
  Department of Computer Engineering \\
  Sharif University of Technology \\
  \texttt{m.poulaei@gmail.com} \\
  \AND
  Kiarash Kiani Feriz \\
  Department of Computer Engineering \\
  Sharif University of Technology \\
  \texttt{kiarashkia138@gmail.com} \\
  \And
  Mohammad Hossein Rohban \\
  Department of Computer Engineering \\
  Sharif University of Technology \\
  \texttt{rohban@sharif.edu} \\
}

\begin{document}

\maketitle

\begin{abstract}
Many organizations fine-tune publicly available pretrained Automatic Speech Recognition (ASR) models and deploy them in black-box settings, assuming limited access provides protection. We show this assumption is fragile: adversarial perturbations crafted on the public base model transfer effectively to fine-tuned target models, severely degrading performance and posing concerns for safety-critical applications. We propose TransferBreaker, a unified fine-tuning framework that suppresses adversarial transfer by integrating Base Adversarial Fine-Tuning, which restricts adversarial training to base-effective perturbations; Latent Jacobian Regularization, which enforces latent-space invariance by suppressing adversarially sensitive directions; and HybridGrad-AFT, which improves robustness against adaptive attacks by interpolating transferable perturbations from base and target gradients. We theoretically justify all components and evaluate TransferBreaker across three languages and four large ASR models, reducing adversarial WER from 92.6 to 27.8. Our code is publicly available at \href{https://github.com/rohban-lab/TransferBreaker}{github.com/rohban-lab/TransferBreaker}.
\end{abstract}
\section{Introduction}\label{sec:introduction}

Deep learning–based Automatic Speech Recognition (ASR) systems have achieved remarkable performance in recent years, yet remain highly vulnerable to adversarial perturbations \cite{pmlr-v97-qin19a}, which are small, carefully crafted input modifications that can induce incorrect or malicious predictions. This vulnerability raises serious trustworthiness concerns, particularly in safety-critical ASR applications such as in-vehicle voice interfaces and large-scale speech transcription services \cite{Sun2024}.

\begin{figure}[t]
        \centering
        \includegraphics[width=\linewidth]{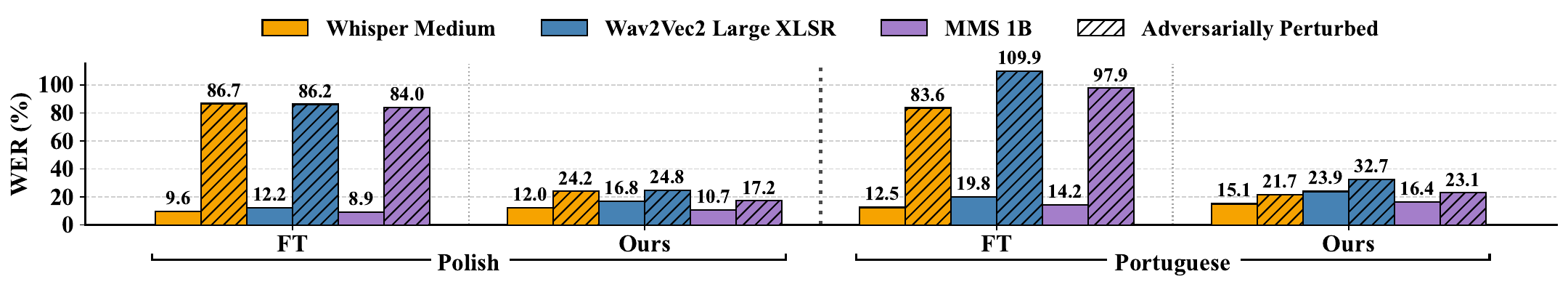}
    \caption{\textbf{Adversarial Transferability from Base to Fine-Tuned ASR Models.}
Fine-tuned (FT) ASR models evaluated across multiple languages and backbones exhibit a sharp increase in Word Error Rate (WER) on adversarial examples optimized against the public base model, relative to clean inputs, revealing a critical transferability vulnerability. In contrast, our proposed framework, \textit{TransferBreaker}, effectively suppresses this transferability and preserves robustness against base-model-derived attacks. \textit{Note:} WER can exceed 100\% due to insertion/deletion errors; see Appendix~\ref{Appendix:Evaluation_Metrics}.}
        \label{fig:motivation_figure}
\vspace{-2pt}
\end{figure}

In many ASR development pipelines, publicly available pretrained models, such as Whisper \cite{whisper}, are fine-tuned to leverage their strong multilingual pretraining and proven architectures. The resulting fine-tuned models are typically deployed as private black-box systems, where restricted access to parameters and gradients is assumed to provide a layer of protection. However, we expose a critical vulnerability that undermines this assumption in practice. As Fig.~\ref{fig:motivation_figure} illustrates, adversarial perturbations crafted with white-box access to the base model transfer effectively to the fine-tuned target. Consequently, attackers with access only to the base model can achieve effectiveness comparable to white-box attacks on the private model, posing a serious threat and highlighting the need for adversarially robust fine-tuning in ASR deployments.

The most effective baseline defense against adversarial attacks is adversarial fine-tuning (AFT) \cite{madry2018towards}, originally proposed for classification, which augments training data with adversarial examples generated by maximizing the model’s loss with respect to the input. As shown in Table~\ref{tab:main_results}, Standard-AFT improves robustness; however, the resulting Word Error Rate (WER) under adversarial evaluation remains prohibitively high for practical ASR deployment.

This limitation arises from two factors. First, ASR is inherently more complex than standard classification; for example, autoregressive models such as Whisper must generate long output sequences, where each token prediction corresponds to a large-vocabulary classification \cite{larg_vocab_long_seq}. Second, Standard-AFT is designed to achieve full white-box robustness and does not exploit the black-box advantage inherent in our setting. Together, these factors cause Standard-AFT to reduce sensitivity to adversarial perturbations across the entire perturbation space for a complex ASR task, leading to an overly broad optimization objective that can excessively constrain the model by enforcing many white-box, worst-case robustness constraints that are unnecessary in our black-box setting.

To address this, we leverage the inductive bias of our transferability threat model: an optimal attacker generates adversarial perturbations against the publicly available base model. An effective defense should therefore minimize the impact of these base-derived attacks on the fine-tuned target model. We propose Base Adversarial Fine-Tuning (Base-AFT), which augments the training data with adversarial examples generated against a fixed base model rather than the continuously updated target model. This restricts adversarial training to the subspace of perturbations that are effective against the base model, rather than the full perturbation space, as theoretically justified in Section \ref{Base_Theory_SubSubSection}.


Based on our observations in Fig.~\ref{fig:LJR_Motivation}, adversarial examples with higher Character Error Rate tend to exhibit larger clean-to-adversarial RMS distance in the target latent space. Since large ASR models may struggle to close this gap using AFT alone, we hypothesize that explicitly enforcing latent-space invariance to adversarial perturbations improves robustness. We therefore propose Latent Jacobian Regularization (LJ-Regularization), which minimizes the squared $\ell_2$ distance between the base model’s latent representation of clean samples and the target model’s latent representation of their adversarial counterparts. Table~\ref{tab:ljr_ablation} confirms the effectiveness of LJ-Regularization, supporting our hypothesis. 
Moreover, by decomposing the LJ-Regularization gradient, Section 4.2.1 shows theoretically how it promotes this invariance through a stop-gradient self-invariance update together with a clean-representation drift correction anchored to the frozen base model, while relating clean-to-adversarial latent displacement to directional latent-Jacobian sensitivity.

Our framework, which combines Base-AFT with LJ-Regularization, achieves strong robustness under our transferability threat model. However, experimental results in Table~\ref{tab:aft_variants} and \ref{tab:component_ablation} indicate that it provides limited protection against adaptive attackers who are fully aware of the defense strategy \cite{Adaptive_Attack}. In this setting, the attacker trains a surrogate model using the same framework on their dataset to approximate the target model, and adversarial perturbations crafted on the surrogate still transfer effectively to the target model despite Base-AFT and LJ-Regularization.

To address this vulnerability, the target model must exhibit a degree of general white-box–level adversarial robustness to resist attacks from surrogate models that closely approximate it. Standard-AFT is effective in white-box settings; however, as discussed earlier, it enforces robustness over the entire perturbation space, which can overly constrain the model and is unnecessary for our threat model. As a result, its robustness against base-model attacks is insufficient.

Motivated by these observations, we propose HybridGrad-AFT, which provides targeted general robustness by restricting the perturbation space to the fundamental and transferable perturbations on which our threat model focuses, rather than the full perturbation space. HybridGrad-AFT augments the training data with adversarial examples perturbed by HybridGrad, constructed by combining base- and target-model gradients to interpolate fundamental and transferable adversarial directions, ensuring transferability between the base and target models. HybridGrad-AFT substantially improves robustness against adaptive attacks, while introducing a trade-off under base-model attacks; LJR partially recovers this loss (Appendix~\ref{appendix:component_ablation}). Appendix~\ref{HybridGrad-AFT} theoretically formalizes this space and shows that HybridGrad-AFT strengthens robustness along fundamental transfer directions. 

We propose \textbf{TransferBreaker}, a fine-tuning framework that suppresses adversarial transferability by integrating Base-AFT, HybridGrad-AFT, and LJ-Regularization into a unified training procedure.\quad\textbf{Contributions.}
(i) we expose a critical vulnerability where white-box access to a base model enables near–white-box attacks on target models, increasing average WER from 16.4 to 92.6; (ii) Base-AFT restricts adversarial training to base-effective perturbations, avoiding the over-constraining behavior of Standard-AFT in complex ASR tasks; (iii) LJ-Regularization enforces latent-space invariance by suppressing adversarially sensitive directions; (iv) we develop HybridGrad-AFT to improve robustness against adaptive attacks by interpolating transferable perturbations from base and target gradients; and (v) we provide theoretical justification and extensive experiments across three languages (Polish, Portuguese, and Arabic) \cite{commonvoice:2020} and four state-of-the-art ASR models: Whisper-large-v3-turbo, wav2vec2-large \cite{Wav2vec}, MMS-1B \cite{MMS}, and Whisper-medium, showing that TransferBreaker reduces average adversarial WER from 92.6 to 27.8 on base-model attacks.

\section{Related Work}
Prior work establishes that ASR systems are vulnerable to adversarial perturbations in both white- and black-box settings \cite{carlini2018audio, pmlr-v97-qin19a, schoenherr2019adversarial}, including cross-model transferability \cite{gao2024transferableadversarialattacksasr, olivier2023transferable, abdullah2021sok}. However, transferability from publicly available pretrained ASR models to their fine-tuned counterparts, particularly in large-scale systems, remains underexplored. We formalize this threat model and propose TransferBreaker to mitigate it. Existing effective defenses primarily rely on adversarial fine-tuning \cite{madry2018towards,HybridSpeech}, which assumes white-box access and can be overly conservative in transfer-based scenarios. Additionally, preprocessing-based defenses \cite{DiffusionDefence, kong2022cleanunet} often fail to mitigate such attacks and may degrade clean performance (see Appendix~\ref{appendix:PreprocessingBaseDefence}). Further discussion is provided in Appendix~\ref{Appendix_Additional_Related_Work}.
\section{Background and Preliminaries}

\begin{figure}[t]
        \centering
        \includegraphics[width=1.0\linewidth]{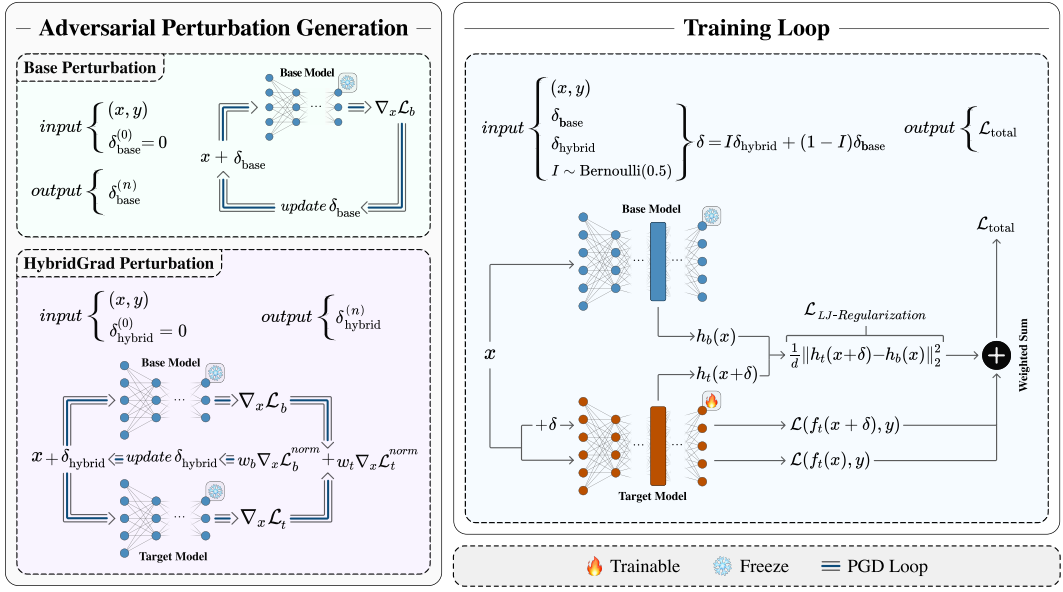}

\caption{\textbf{The TransferBreaker Framework.} \textbf{Left:} Adversarial perturbation generation, including Base Perturbation ($\delta_{\text{base}}$) and HybridGrad Perturbation ($\delta_{\text{hybrid}}$), where the latter combines base and target gradients to interpolate transferable adversarial directions. \textbf{Right:} The training loop using clean and adversarial samples with task loss (CTC or Cross-Entropy) and Latent Jacobian Regularization to enforce latent-space invariance and suppress adversarially sensitive directions.}
\label{fig:main_figure}
\end{figure}


\paragraph{Adversarial Attack on ASR.} 
An adversarial attack modifies an input sample \(x\) with label \(y\) to maximize the loss \(\ell(x; y)\) \cite{yuan2019adversarial, xu2019adversarialattacksdefensesimages}. The resulting sample \(x^*\) is called an adversarial example, and \(x^* - x\) is the adversarial perturbation, typically constrained by an \(l_p\)-norm bound \(\|x - x'\|_p \le \epsilon\) to ensure semantic preservation. Formally,
\(
x^* = \arg \max_{x': \|x - x'\|_p \le \epsilon} \ell(x'; y).
\) Projected Gradient Descent (PGD) \cite{madry2018towards} is a standard first-order method that iteratively updates the input by ascending the loss surface while enforcing the perturbation constraint:
\(
x_0^* = x,\quad
x_{t+1}^* = \Pi_{\mathcal{B}_\epsilon(x)}\!\left(
x_t^* + \alpha \cdot \operatorname{sign}\big(\nabla_x \ell(x_t^*; y)\big)
\right),
\quad
x^* = x_K^*,
\)
where \(\alpha\) is the step size, \(K\) is the number of iterations, and \(\Pi_{\mathcal{B}_\epsilon(x)}(\cdot)\) denotes projection onto the \(l_p\)-ball centered at \(x\) with radius \(\epsilon\).

In ASR, adversarial attacks differ from classification because outputs are variable-length sequences $\hat{y} = (\hat{y}_1, \dots, \hat{y}_N)$, often with large $N$, leading to correlated token-level predictions rather than a single prediction. The adversarial objective $\ell(\cdot)$ is therefore defined over the full sequence (e.g., CTC or cross-entropy) to induce transcription errors. Moreover, modern ASR operates in large-vocabulary settings (e.g., $>50{,}000$ tokens in Whisper), unlike classification tasks with far fewer classes. To ensure nearly imperceptible perturbations, instead of a fixed $l_p$-norm constraint, we enforce $\mathrm{SNR}(x, x^*) \geq 27~\mathrm{dB}$, expressed as the signal-dependent constraint {\scriptsize $\|x - x^*\|_{\infty} \leq \sqrt{\frac{P(x)}{10^{27/10}}}$}, where $P(x)$ denotes the mean signal power. Further perceptual analysis is provided in Appendix~\ref{appendix:perceptual_analysis}.

\textbf{Threat Model.}\label{sec:threat_model}
Given an input audio waveform $x$ and its ground-truth transcription $y$, we consider an adversary whose goal is to generate a nearly imperceptible, untargeted adversarial perturbation $\delta$ to degrade a privately fine-tuned ASR system, without ever querying $f_t$ during perturbation construction (a \emph{zero-query transfer} setting). The adversary has white-box access to a publicly available base model $f_b(\cdot)$ (e.g., Whisper), including its parameters and gradients, but no access to the fine-tuned target model $f_t(\cdot)$, which is treated as a black box derived from the base model.

The perturbation $\delta$ is optimized on $f_b$ under perceptual constraints to maximize the training loss for a given audio sample (e.g., CTC or seq2seq cross-entropy), leading to increased transcription errors measured by WER. Formally, \(\mathrm{WER}(f_b(x+\delta), y) \gg \mathrm{WER}(f_b(x), y),\) where $f_b(x+\delta) = (\hat{y}_1,\dots,\hat{y}_M)$ and $y = (y_1,\dots,y_N)$. Since ASR outputs are variable-length sequences, the optimization induces errors across the entire sequence rather than at a single token position. Leveraging the transferability of adversarial examples across fine-tuned models, the same perturbation $\delta$ is applied to the private target model $f_t$ without adaptation. The attack is considered successful if it significantly degrades transcription quality, i.e.,
\(\mathrm{WER}\big(f_t(x + \delta), y\big) \gg \mathrm{WER}\big(f_t(x), y\big).\)

This threat model reflects realistic deployments where organizations, especially for mid- and low-resource languages, fine-tune public ASR models for their strong performance. Although the target model is a black box, we show perturbations crafted on the base model transfer effectively, often approaching white-box strength, and this persists even under large-scale fine-tuning (up to 1,000 hours, Appendix~\ref{appendix:FT_on_more_data}), indicating a fundamental vulnerability that challenges the assumed security of black-box deployment and motivates robust fine-tuning. Section~\ref{sec:adaptive_thread_model} extends this to a defense-aware zero-query adversary that knows the TransferBreaker procedure and public base model but has no access to the target model's parameters, gradients, or training data. Using its own data, the attacker trains a surrogate $f_s$ from the base model and transfers perturbations crafted on $f_s$ to $f_t$. We justify our motivation and scope of untargeted attacks in Appendix~\ref{appendix:untargeted_scope}.

\section{Methodology}
\textbf{Overview.} Many privately fine-tuned ASR models \(f_t\) deployed by organizations are highly vulnerable to adversarial perturbations crafted on the publicly available base models \(f_b\) on which they are fine-tuned. We propose TransferBreaker, an adversarial fine-tuning framework that suppresses transferability from base to target models. TransferBreaker generates adversarial examples using two mechanisms: base-model gradients and HybridGrad, which jointly leverages base- and target-model gradients. Base-model gradients exploit alignment with attacker objectives to reduce transferability, while HybridGrad improves robustness to adaptive attacks by focusing perturbations on transferable directions rather than the entire perturbation space. We further introduce Latent Jacobian Regularization, which constrains adversarial latent displacement using a frozen base-model reference, strengthening transferability suppression. We provide theoretical justification for base-model adversarial examples, HybridGrad adversarial examples, and LJ-Regularization. The following subsections describe the main components.

\subsection{Base Adversarial Fine-Tuning}\label{section:Base-Adversarial-Fine-Tuning}
Let $f_b$ denote the public base model and $f_t$ the privately fine-tuned target model. We define their training losses as
{
\setlength{\abovedisplayskip}{4pt}
\setlength{\belowdisplayskip}{4pt}
\begin{equation}
\mathcal{L}_b(x, y) \triangleq \mathcal{L}(f_b(x), y), \quad
\mathcal{L}_t(x,y) \triangleq \mathcal{L}(f_t(x), y) \, .
\end{equation}}
In our threat model, we assume that the attacker has access to the base model but not the final target model. Accordingly, the attacker attempts to compute $\delta_{\mathrm{base}}$ such that
{
\setlength{\abovedisplayskip}{3pt}
\setlength{\belowdisplayskip}{3pt}
\begin{equation}\label{Eq_Attacker_Objective_Our_Threat_Model}
\delta_{\mathrm{base}} = \arg\max_{\|\delta\|_p \le \epsilon} \mathcal{L}_b(x+\delta, y) \, .
\end{equation}}
An optimal defender should minimize the impact of this base-derived attack on the target model
{
\setlength{\abovedisplayskip}{3pt}
\setlength{\belowdisplayskip}{0pt}
\setlength{\abovedisplayshortskip}{3pt}
\setlength{\belowdisplayshortskip}{0pt}
\begin{equation}\label{Eq_Base_AT_Optimal_Objective}
\min_{\theta_t}
\mathbb{E}_{(x,y)} \left[ \mathcal{L}_t(x+\delta_{\mathrm{base}}, y) \right] \, .
\end{equation}}

We refer to this strategy as \emph{base adversarial fine-tuning (Base-AFT)}, in which the training data are augmented with adversarial examples generated against the base model using a momentum-based PGD attack \cite{dong2018momentum}. The procedure is detailed in Algorithm~\ref{alg:delta_base} (Appendix~\ref{appendix:algorithms_jacobian}).

\paragraph{Empirical motivation.}
Our experiments (Table~\ref{tab:base_ablation}) show that Base-AFT is substantially more effective than Standard-AFT against attacks crafted on the base model. While prior work identifies Standard-AFT (i.e., augmenting training data with adversarial examples generated against the current model) as the strongest defense in white-box settings, our results show that it remains substantially less effective under our threat model, with WER remaining high under transferred attacks (see Appendix~\ref{appendix:std-aft-across-epochs} for further analysis).

We attribute this limitation to the inherent complexity of ASR. Unlike classification, ASR involves long sequence generation with large-vocabulary predictions at each step. Standard-AFT therefore attempts to reduce sensitivity to adversarial perturbations across the entire perturbation space for a complex sequence-generation task, leading to an overly broad optimization objective that can excessively constrain the model. In contrast, Equations~\eqref{Eq_Attacker_Objective_Our_Threat_Model} and \eqref{Eq_Base_AT_Optimal_Objective} explain why Base-AFT is optimal. By aligning the training-time adversarial objective with the attacker’s objective, namely perturbations generated on the publicly available base model, Base-AFT directly targets the subspace responsible for transferability. Moreover, fixing adversarial generation to the static base model reduces optimization complexity while focusing robustness on the perturbation directions that matter most.


In Section~\ref{Base_Theory_SubSubSection}, we formalize this intuition and theoretically show that Base-AFT concentrates robustness on a structured subspace of directions to which the base model is most vulnerable, rather than the entire perturbation space.



\subsubsection{Theoretical Analysis of Base-AFT versus Standard-AFT}\label{Base_Theory_SubSubSection}
\textbf{Mechanism of Base-AFT.} Using a Taylor expansion (see Appendix~\ref{appendix:why_taylor_expansion} for justification), for small $\epsilon$ we have
{
\setlength{\abovedisplayskip}{3pt}
\setlength{\belowdisplayskip}{0pt}
\begin{equation}
\mathcal{L}_b(x+\delta, y) \approx \mathcal{L}_b(x,y) + \nabla_x \mathcal{L}_b(x,y) \cdot \delta \, .
\end{equation}}%
So
{
\setlength{\abovedisplayskip}{0pt}
\setlength{\belowdisplayskip}{3pt}
\begin{equation}
\begin{gathered}
\delta_{\mathrm{base}}
= \arg\max_{\|\delta\|_p\le\epsilon} \mathcal{L}_b(x+\delta,y)
\approx \arg\max_{\|\delta\|_p\le\epsilon}
\big( \mathcal{L}_b(x,y) + \nabla_x \mathcal{L}_b(x,y)\cdot\delta \big)
= \\
\arg\max_{\|\delta\|_p \le \epsilon}
\nabla_x \mathcal{L}_b(x,y) \cdot \delta \, .
\end{gathered}
\end{equation}
}

Solving this optimization problem gives $\delta_{\mathrm{base}} \approx \epsilon\,u_b(x)$, where $u_b(x)$ is the unit $\ell_p$-norm direction that maximizes the inner product with the base gradient $\nabla_x \mathcal{L}_b(x,y)$. Using the same first-order approximation for the target loss and substituting $\delta_{\mathrm{base}}$, the defender's objective becomes
{
\setlength{\abovedisplayskip}{3pt}
\setlength{\belowdisplayskip}{3pt}
\setlength{\abovedisplayshortskip}{3pt}
\setlength{\belowdisplayshortskip}{3pt}
\begin{equation}
\begin{gathered}
\min_{\theta_t} \mathbb{E}_{(x,y)} \bigl[ \mathcal{L}_t(x+\delta_{\mathrm{base}}, y) \bigr]
\;\approx\;
\min_{\theta_t} \mathbb{E}_{(x,y)} \bigl[ \mathcal{L}_t(x,y) + \nabla_x \mathcal{L}_t(x,y)\cdot \epsilon\, u_b(x) \bigr]
=
\\[0pt]
\min_{\theta_t} \Bigl( \mathbb{E}_{(x,y)} \bigl[ \mathcal{L}_t(x,y) \bigr] + \epsilon\, \mathbb{E}_{(x,y)} \bigl[ \left\langle \nabla_x \mathcal{L}_t(x,y),\,u_b(x)\right\rangle \bigr] \Bigr) \, .
\end{gathered}
\end{equation}
}

\textbf{Interpretation.}
Let $\mathcal{U}_b(x)=\operatorname{span}\{u_b(x)\}$ denote the adversarial subspace identified by the base model. Base-AFT seeks to maintain clean accuracy while strategically reducing the target model's loss response along the adversarial direction $u_b(x)$ within $\mathcal{U}_b(x)$, rather than attempting to cover the entire perturbation space.

\textbf{Mechanism of Standard-AFT.} In standard adversarial fine-tuning, we optimize the objective
\begin{equation}
\min_{\theta_t}
\mathbb{E}_{(x,y)} \left[
\max_{\|\delta\|_p \le \epsilon} \mathcal{L}_t(x+\delta,y)
\right] \, .
\end{equation}
Based on Taylor expansion, for small $\epsilon$ we have
{
\setlength{\abovedisplayskip}{3pt}
\setlength{\belowdisplayskip}{3pt}
\begin{equation}
\mathcal{L}_t(x+\delta,y) \approx \mathcal{L}_t(x,y) + \nabla_x \mathcal{L}_t(x,y) \cdot \delta \, .
\end{equation}}
So
{
\setlength{\abovedisplayskip}{3pt}
\setlength{\belowdisplayskip}{3pt}
\begin{equation}
\max_{\|\delta\|_p \le \epsilon} \mathcal{L}_t(x+\delta,y)
\approx
\mathcal{L}_t(x,y) +
\max_{\|\delta\|_p \le \epsilon}
\nabla_x \mathcal{L}_t(x,y) \cdot \delta \, .
\end{equation}}
Solving this optimization problem gives
{
\setlength{\abovedisplayskip}{3pt}
\setlength{\belowdisplayskip}{3pt}
\begin{equation}
\max_{\|\delta\|_p \le \epsilon} \mathcal{L}_t(x+\delta,y)
\approx
\mathcal{L}_t(x,y)
+
\epsilon \ \| \nabla_x \mathcal{L}_t(x,y) \|_q \, ,
\end{equation}}
where $\frac{1}{p} + \frac{1}{q} = 1$. Substituting this approximation into the defender’s problem yields
\begin{equation}
\operatorname*{min}_{\theta_t}\,
\mathbb{E}_{(x,y)}
\Big[
\operatorname*{max}_{\lVert\delta\rVert_p\le\epsilon}
\mathcal{L}_t(x+\delta,y)
\Big]
\;\approx\;
\operatorname*{min}_{\theta_t}\,
\Big(
\mathbb{E}_{(x,y)}[\mathcal{L}_t(x,y)]
+\epsilon\,\mathbb{E}_{(x,y)}[\lVert\nabla_x \mathcal{L}_t(x,y)\rVert_q]
\Big) \, .
\end{equation}
\textbf{Comparison.}
Standard-AFT seeks to maintain clean accuracy while reducing the target model's sensitivity $(\nabla_x \mathcal{L}_t(x,y))$ over the entire perturbation space. This is a broad optimization objective and can overly constrain the model. In contrast, Base-AFT focuses robustness on the attacker-relevant subspace induced by the base model, yielding stronger protection against transferred attacks.


\subsection{Latent Jacobian Regularization}
\begin{wrapfigure}{r}{0.55\textwidth}
    \vspace{-3\baselineskip}
    \centering
    \includegraphics[width=\linewidth]{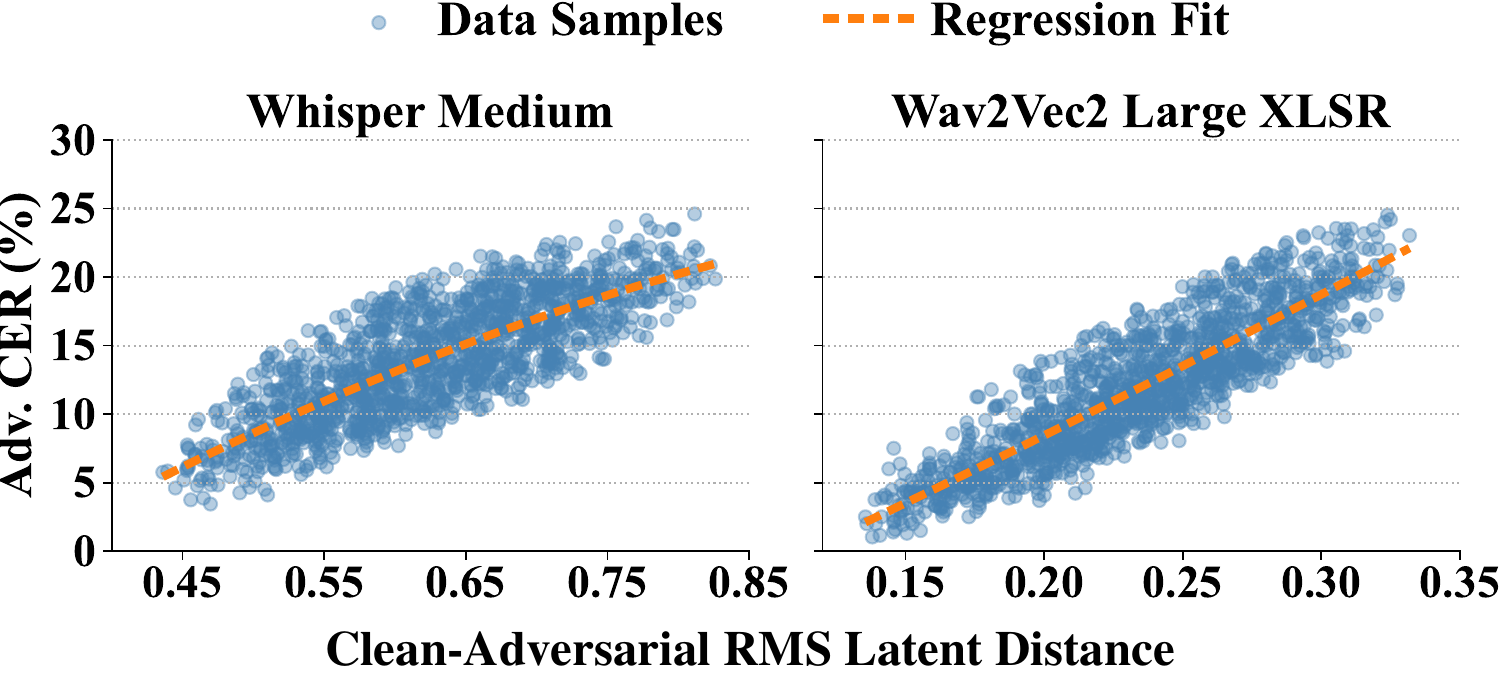}
    \caption{\textbf{Motivation for Latent Jacobian Regularization.} For Base-AFT target models on Polish, adversarial CER is positively associated with clean-to-adversarial RMS latent distance. Each point corresponds to an evaluation utterance, and the dashed lines show regression fits. This association motivates explicitly constraining adversarial latent displacement during fine-tuning.}
    \label{fig:LJR_Motivation}
    \vspace{0.0\baselineskip}
\end{wrapfigure}

Based on our observations on the evaluation set (see Fig.~\ref{fig:LJR_Motivation}), we find that adversarial examples with higher CER also tend to exhibit larger clean-to-adversarial RMS distance in the target latent space. This suggests that AFT alone may not sufficiently constrain adversarial latent displacement. We therefore hypothesize that explicitly constraining latent representations under adversarial perturbations, in addition to AFT, can yield stronger robustness.

\begin{table*}[t]
\caption{Main results comparing WER/CER under clean and transferred adversarial settings across four ASR backbones. Base model and its variants are trained under three regimes: fine-tuning (FT), standard adversarial fine-tuning (Std-AFT), and TransferBreaker (Ours), using three languages from Common Voice~24. ``Base-Adv'' denotes adversarial examples crafted on the base model and transferred to the corresponding target models. Lower is better. \textit{Note:} WER can exceed 100\% due to insertion/deletion errors; see Appendix~\ref{Appendix:Evaluation_Metrics}.}

\resizebox{\linewidth}{!}{%
\centering
\begin{tabular}{
  BBB c
  c c c >{\columncolor[HTML]{e5e5ff}}c
  c c c >{\columncolor[HTML]{e5e5ff}}c
  c c c >{\columncolor[HTML]{e5e5ff}}c
  c c c >{\columncolor[HTML]{e5e5ff}}c
}

\specialrule{1.5pt}{\aboverulesep}{\belowrulesep}

 &  &  &
 & \multicolumn{4}{c}{\textbf{Whisper Medium}}
 & \multicolumn{4}{c}{\textbf{Wav2Vec2 Large XLSR}}
 & \multicolumn{4}{c}{\textbf{MMS 1B}}
 & \multicolumn{4}{c}{\textbf{Whisper Large V3 Turbo}} \\
\cmidrule(lr){5-8}\cmidrule(lr){9-12}\cmidrule(lr){13-16}\cmidrule(lr){17-20}
\centering\parbox{2.1cm}{\centering \textbf{Language}} & \textbf{Setup} & \textbf{Metric}
 &
 & \textbf{Base} & \textbf{FT} & \textbf{Std-AFT} & \textbf{Ours}
 & \textbf{Base} & \textbf{FT} & \textbf{Std-AFT} & \textbf{Ours}
 & \textbf{Base} & \textbf{FT} & \textbf{Std-AFT} & \textbf{Ours}
 & \textbf{Base} & \textbf{FT} & \textbf{Std-AFT} & \textbf{Ours} \\

\specialrule{1.5pt}{\aboverulesep}{\belowrulesep}

\multirow{4}{*}{\centering\parbox{2.1cm}{\centering Polish}}
 & \multirow{2}{*}{\color{CG}Clean} & \color{CG}WER &
 & \color{CG}15.3 &  \color{CG}9.6 & \color{CG}12.9 & \color{CG}12.0
 & \color{CG}18.8 & \color{CG}12.2 & \color{CG}17.3 & \color{CG}16.8
 & \color{CG}12.9 &  \color{CG}8.9 & \color{CG}12.4 & \color{CG}10.7
 & \color{CG}13.4 &  \color{CG}9.6 & \color{CG}10.8 & \color{CG}11.8 \\
 &  & \color{CG}CER &
 &  \color{CG}5.9 &  \color{CG}2.7 &  \color{CG}4.2 &  \color{CG}3.8
 &  \color{CG}4.4 &  \color{CG}3.0 &  \color{CG}4.5 &  \color{CG}4.2
 &  \color{CG}3.1 &  \color{CG}2.1 &  \color{CG}3.2 &  \color{CG}2.6
 &  \color{CG}4.3 &  \color{CG}2.6 &  \color{CG}3.3 &  \color{CG}3.7 \\

\cmidrule(lr){2-3} \cmidrule(lr){5-8}\cmidrule(lr){9-12}\cmidrule(lr){13-16}\cmidrule(lr){17-20}

 & \multirow{2}{*}{Base-Adv} & WER &
 & 118.1 &  86.7 & 44.0 & \textbf{24.2}
 & 102.0 &  86.2 & 33.7 & \textbf{24.8}
 & 104.7 &  84.0 & 28.7 & \textbf{17.2}
 & 128.0 &  68.4 & 32.2 & \textbf{20.8} \\
 &  & CER &
 &  83.2 &  46.0 & 22.5 &  \textbf{9.7}
 &  54.7 &  46.2 & 10.3 &  \textbf{6.9}
 &  63.7 &  44.3 &  8.8 &  \textbf{4.8}
 &  78.8 &  32.8 & 14.2 &  \textbf{7.8} \\

\specialrule{1.5pt}{\aboverulesep}{\belowrulesep}

\multirow{4}{*}{\centering\parbox{2.1cm}{\centering Portuguese}}
 & \multirow{2}{*}{\color{CG}Clean} & \color{CG}WER &
 & \color{CG}22.2 & \color{CG}12.5 & \color{CG}16.1 & \color{CG}15.1
 & \color{CG}30.0 & \color{CG}19.8 & \color{CG}26.4 & \color{CG}23.9
 & \color{CG}19.8 & \color{CG}14.2 & \color{CG}18.2 & \color{CG}16.4
 & \color{CG}16.9 & \color{CG}12.6 & \color{CG}12.9 & \color{CG}12.7 \\
 &  & \color{CG}CER &
 &  \color{CG}7.0 &  \color{CG}3.7 &  \color{CG}5.5 &  \color{CG}4.8
 &  \color{CG}7.8 &  \color{CG}5.2 &  \color{CG}7.7 &  \color{CG}6.7
 &  \color{CG}5.2 &  \color{CG}3.7 &  \color{CG}4.9 &  \color{CG}4.3
 &  \color{CG}5.8 &  \color{CG}4.1 &  \color{CG}4.3 &  \color{CG}4.1 \\

\cmidrule(lr){2-3} \cmidrule(lr){5-8}\cmidrule(lr){9-12}\cmidrule(lr){13-16}\cmidrule(lr){17-20}

 & \multirow{2}{*}{Base-Adv} & WER &
 & 147.9 &  83.6 & 42.9 & \textbf{21.7}
 & 117.2 & 109.9 & 41.9 & \textbf{32.7}
 & 171.0 &  97.9 & 34.5 & \textbf{23.1}
 & 158.6 &  75.8 & 45.3 & \textbf{19.3} \\
 &  & CER &
 &  86.7 &  44.7 & 21.1 &  \textbf{8.2}
 &  62.6 &  59.3 & 14.1 & \textbf{10.7}
 &  84.9 &  52.5 & 11.5 &  \textbf{7.1}
 &  89.1 &  42.0 & 18.6 &  \textbf{7.2} \\

\specialrule{1.5pt}{\aboverulesep}{\belowrulesep}

\multirow{4}{*}{\centering\parbox{2.1cm}{\centering Arabic}}
 & \multirow{2}{*}{\color{CG}Clean} & \color{CG}WER &
 & \color{CG}38.4 & \color{CG}24.7 & \color{CG}28.1 & \color{CG}26.2
 & \color{CG}35.0 & \color{CG}22.9 & \color{CG}28.2 & \color{CG}26.4
 & \color{CG}34.1 & \color{CG}24.7 & \color{CG}31.0 & \color{CG}28.5
 & \color{CG}37.2 & \color{CG}24.9 & \color{CG}28.8 & \color{CG}26.3 \\
 &  & \color{CG}CER &
 & \color{CG}11.0 &  \color{CG}6.4 &  \color{CG}8.5 &  \color{CG}8.3
 &  \color{CG}8.3 &  \color{CG}5.4 &  \color{CG}6.8 &  \color{CG}6.2
 &  \color{CG}8.1 &  \color{CG}5.9 &  \color{CG}7.4 &  \color{CG}6.6
 &  \color{CG}9.7 &  \color{CG}6.2 &  \color{CG}7.8 &  \color{CG}7.2 \\

\cmidrule(lr){2-3} \cmidrule(lr){5-8}\cmidrule(lr){9-12}\cmidrule(lr){13-16}\cmidrule(lr){17-20}

 & \multirow{2}{*}{Base-Adv} & WER &
 & 164.2 & 110.0 & 58.4 & \textbf{36.5}
 & 124.9 & 105.1 & 48.5 & \textbf{36.8}
 & 136.4 &  97.3 & 50.9 & \textbf{38.9}
 & 158.4 & 106.1 & 52.6 & \textbf{38.0} \\
 &  & CER &
 &  85.2 &  56.5 & 27.1 & \textbf{12.2}
 &  67.3 &  53.3 & 14.3 & \textbf{10.6}
 &  66.5 &  48.4 & 14.1 & \textbf{10.7}
 &  79.7 &  53.6 & 20.5 & \textbf{11.8} \\

\specialrule{1.5pt}{\aboverulesep}{\belowrulesep}

\end{tabular}
}
\label{tab:main_results}
\end{table*}

This observation motivates explicitly controlling adversarial latent displacement. We further find that using the frozen base model's clean representation as the reference is better aligned with our base-to-target transfer objective and yields stronger robustness than using the target model's own clean representation as the reference (see Appendix~\ref{appendix:ljr_anchor_ablation}). We therefore propose \emph{Latent Jacobian Regularization (LJ-Regularization)}, which penalizes the squared $\ell_2$ distance between the frozen base model's latent representation $h_b(x)$ of a clean sample and the target model's latent representation $h_t(x+\delta)$ of its adversarial counterpart. Let $x$ denote an audio input, $\delta$ an adversarial perturbation, and $d$ the number of latent elements being compared. We define
{
\setlength{\abovedisplayskip}{1pt}
\setlength{\belowdisplayskip}{3pt}
\setlength{\abovedisplayshortskip}{1pt}
\setlength{\belowdisplayshortskip}{3pt}
\begin{equation}\label{LJR_main_formula}
\mathcal{L}_{\mathrm{LJ-Regularization}} = \frac{1}{d}\,\big\| h_t(x+\delta) - h_b(x) \big\|_2^2 \, .
\end{equation}}
Moreover, using the frozen base model as a fixed reference avoids the trivial representation-collapse solution that can arise when optimizing a squared distance between two outputs of the same trainable model \cite{grill2020bootstrap,chen2021exploring}.

In Section~\ref{LJR_theory}, we formalize LJ-Regularization and characterize its connection to latent invariance. Specifically, we show that its outer update consists of a stop-gradient self-invariance update together with a clean-representation drift correction anchored to the frozen base model. We further relate the clean-to-adversarial latent displacement to directional latent-Jacobian sensitivity.

\subsubsection{Theoretical Analysis of LJ-Regularization}\label{LJR_theory}

We characterize LJ-Regularization relative to stop-gradient self-invariance without assuming that the clean target representation $h_t(x)$ is close to the base representation $h_b(x)$. Define the clean representation drift and the clean-to-adversarial target displacement as
{
\setlength{\abovedisplayskip}{3pt}
\setlength{\belowdisplayskip}{3pt}
\begin{equation}
 g(x) \triangleq h_t(x)-h_b(x),
 \qquad
 \Delta(x,\delta) \triangleq h_t(x+\delta)-h_t(x).
 \label{eq:ljr_drift_delta}
\end{equation}}%
Since $h_t(x+\delta)-h_b(x)=\Delta(x,\delta)+g(x)$, the LJ-Regularization objective in Eq.~\ref{LJR_main_formula} can be written exactly as
{
\setlength{\abovedisplayskip}{1pt}
\setlength{\belowdisplayskip}{3pt}
\begin{equation}
 d\mathcal{L}_{\mathrm{LJ-Regularization}}
 = \|\Delta\|_2^2 + \|g(x)\|_2^2 + 2\langle \Delta,g(x)\rangle.
 \label{eq:ljr_objective_decomposition}
\end{equation}}%
For comparison, define the stop-gradient self-invariance objective
{
\setlength{\abovedisplayskip}{1pt}
\setlength{\belowdisplayskip}{3pt}
\begin{equation}
 \mathcal{L}_{\mathrm{self}}
 = \frac{1}{d}\left\|h_t(x+\delta)-\operatorname{sg}[h_t(x)]\right\|_2^2.
 \label{eq:self_ljr}
\end{equation}}%
Its forward value is $d\mathcal{L}_{\mathrm{self}}=\|\Delta\|_2^2$. Equation~\ref{eq:ljr_objective_decomposition} is an algebraic decomposition of the frozen-base objective; its three terms should not be interpreted as independently optimized regularizers.

At the level of the outer training update, the distinction between the two objectives is exact. Let $\theta$ denote the target-model parameters, and let $J_\theta h_t(x+\delta)$ denote the Jacobian of the target latent representation with respect to $\theta$. The base model is frozen, and the generated perturbation $\delta$ is detached during the outer parameter update. Therefore,
{
\setlength{\abovedisplayskip}{1pt}
\setlength{\belowdisplayskip}{3pt}
\begin{equation}
\nabla_\theta \mathcal{L}_{\mathrm{LJ-Regularization}}
=
\nabla_\theta \mathcal{L}_{\mathrm{self}}
+
\frac{2}{d}J_\theta h_t(x+\delta)^\top g(x).
\label{eq:ljr_gradient_identity}
\end{equation}}%
At initialization, $h_t=h_b$, so $g(x)=0$ and the two updates coincide. During fine-tuning, $g(x)$ generally becomes nonzero. Equation~\ref{eq:ljr_gradient_identity} shows that the LJ-Regularization outer update equals the stop-gradient self-invariance update plus a correction determined by the clean representation drift. The self objective aligns $h_t(x+\delta)$ with the current clean target representation $h_t(x)$, whereas LJ-Regularization aligns $h_t(x+\delta)$ with the fixed clean base representation $h_b(x)$. Thus, the frozen-base objective retains explicit dependence on the public base model throughout fine-tuning, matching the base-to-target transfer setting studied here.

The clean-to-adversarial displacement $\Delta(x,\delta)$ also has a local latent-Jacobian interpretation. For sufficiently small $\delta$,
{
\setlength{\abovedisplayskip}{1pt}
\setlength{\belowdisplayskip}{3pt}
\begin{equation}
\Delta(x,\delta)
=
J_{h_t}(x)\delta + r(x,\delta).
\label{eq:delta_jacobian_local}
\end{equation}}%
Here, $J_{h_t}(x)$ is the Jacobian of $h_t$ with respect to the input and $r(x,\delta)$ is the higher-order remainder. When the first-order approximation is informative, $\|\Delta(x,\delta)\|_2^2 \approx \|J_{h_t}(x)\delta\|_2^2$. This relation connects clean-to-adversarial latent displacement to directional latent-Jacobian sensitivity. Equation~\ref{eq:ljr_gradient_identity} then shows how the frozen-base outer update augments the corresponding self-invariance update with the drift-dependent base-reference correction. Appendix~\ref{appendix:ljr_anchor_ablation} directly compares the two anchors.

\subsection{Adaptive Adversarial Attacks Against Our Defense}

Although Base-AFT combined with LJ-Regularization effectively suppresses the transferability of adversarial perturbations from the base model, we observe that it provides only limited protection against adaptive attackers who are fully aware of our defense strategy.

\textbf{Adaptive Attack Threat Model.}\label{sec:adaptive_thread_model}
We assume an adaptive attacker with full knowledge of our defense framework, including Base-AFT, LJ-Regularization, and HybridGrad (described later in this section). Using its own dataset, the attacker applies the same TransferBreaker procedure to train a surrogate model \(f_s(\cdot)\) in an attempt to approximate our privately fine-tuned target model \(f_t(\cdot)\). Adversarial perturbations \(\delta\) crafted against \(f_s\) are observed to transfer effectively to \(f_t\), despite the use of Base-AFT and LJ-Regularization during training.


\begin{table}[t]
\centering
\caption{Comparison of Base-AFT, Base+Standard-AFT, and TransferBreaker under clean settings, base-model adversarial attacks (Base-Adv), and surrogate-based adaptive adversarial attacks (Surrogate-Adv) for Polish and Portuguese across ASR backbones. Lower is better.}
\resizebox{1.0\linewidth}{!}{%
\begin{tabular}{
  BBB c
  c c >{\columncolor[HTML]{e5e5ff}}c
  c c >{\columncolor[HTML]{e5e5ff}}c
  c c >{\columncolor[HTML]{e5e5ff}}c
}

\specialrule{1.5pt}{\aboverulesep}{\belowrulesep}

 &  &  &
 & \multicolumn{3}{c}{\textbf{Whisper Medium}}
 & \multicolumn{3}{c}{\textbf{Wav2Vec2 Large XLSR}}
 & \multicolumn{3}{c}{\textbf{MMS 1B}} \\
\cmidrule(lr){5-7}\cmidrule(lr){8-10}\cmidrule(lr){11-13}
Language & Setup & Metric
 &
 & \textbf{Base-AFT} & \textbf{Base+Std-AFT} & \textbf{Ours}
 & \textbf{Base-AFT} & \textbf{Base+Std-AFT} & \textbf{Ours}
 & \textbf{Base-AFT} & \textbf{Base+Std-AFT} & \textbf{Ours} \\

\specialrule{1.5pt}{\aboverulesep}{\belowrulesep}

\multicolumn{1}{c}{\multirow{6}{*}{\textbf{Polish}}}
 & \multirow{2}{*}{\color{CG}Clean} & \color{CG}WER &  & \color{CG}11.0 & \color{CG}13.1 & \color{CG}12.0 & \color{CG}15.9 & \color{CG}16.7 & \color{CG}16.8 & \color{CG}10.6 & \color{CG}11.0 & \color{CG}10.7 \\
 &  & \color{CG}CER &  & \color{CG}3.4 & \color{CG}4.2 & \color{CG}3.8 & \color{CG}3.9 & \color{CG}4.2 & \color{CG}4.2 & \color{CG}2.6 & \color{CG}2.6 & \color{CG}2.6 \\

\cmidrule(lr){2-3}\cmidrule(lr){5-7}\cmidrule(lr){8-10}\cmidrule(lr){11-13}

 & \multirow{2}{*}{Base-Adv} & WER &  & \textbf{22.4} & 30.2 & 24.2 & \textbf{24.0} & 28.9 & 24.8 & 17.2 & 21.1 & \textbf{17.2} \\
 &  & CER &  & \textbf{8.3} & 13.2 & 9.7 & 6.9 & 8.5 & \textbf{6.9} & 4.9 & 6.2 & \textbf{4.8} \\

\cmidrule(lr){2-3}\cmidrule(lr){5-7}\cmidrule(lr){8-10}\cmidrule(lr){11-13}

 & \multirow{2}{*}{Surrogate-Adv} & WER &  & 57.6 & 46.8 & \textbf{44.1} & 54.2 & 44.7 & \textbf{40.7} & 48.6 & 38.0 & \textbf{37.2} \\
 &  & CER &  & 29.6 & 22.7 & \textbf{20.5} & 23.5 & 16.6 & \textbf{15.3} & 19.7 & 13.5 & \textbf{13.3} \\

\specialrule{1.5pt}{\aboverulesep}{\belowrulesep}

\multicolumn{1}{c}{\multirow{6}{*}{\textbf{Portuguese}}}
 & \multirow{2}{*}{\color{CG}Clean} & \color{CG}WER &  & \color{CG}14.2 & \color{CG}16.2 & \color{CG}15.1 & \color{CG}23.2 & \color{CG}24.6 & \color{CG}23.9 & \color{CG}16.2 & \color{CG}17.5 & \color{CG}16.4 \\
 &  & \color{CG}CER &  & \color{CG}4.4 & \color{CG}5.5 & \color{CG}4.8 & \color{CG}6.6 & \color{CG}7.1 & \color{CG}6.7 & \color{CG}4.1 & \color{CG}4.6 & \color{CG}4.3 \\

\cmidrule(lr){2-3}\cmidrule(lr){5-7}\cmidrule(lr){8-10}\cmidrule(lr){11-13}

 & \multirow{2}{*}{Base-Adv} & WER &  & \textbf{19.6} & 27.8 & 21.7 & 32.7 & 36.7 & \textbf{32.7} & \textbf{22.2} & 28.8 & 23.1 \\
 &  & CER &  & \textbf{6.8} & 11.9 & 8.2 & \textbf{10.6} & 12.1 & 10.7 & \textbf{6.6} & 9.3 & 7.1 \\

\cmidrule(lr){2-3}\cmidrule(lr){5-7}\cmidrule(lr){8-10}\cmidrule(lr){11-13}

 & \multirow{2}{*}{Surrogate-Adv} & WER &  & 56.4 & 46.3 & \textbf{44.7} & 62.8 & 51.9 & \textbf{48.9} & 60.3 & 43.7 & \textbf{42.8} \\
 &  & CER &  & 27.8 & 22.0 & \textbf{20.8} & 27.7 & 19.8 & \textbf{18.1} & 27.6 & 17.5 & \textbf{17.0} \\

\specialrule{1.5pt}{\aboverulesep}{\belowrulesep}

\end{tabular}
}
\label{tab:aft_variants}
\end{table}

\textbf{HybridGrad-AFT.}
The success of adaptive attacks indicates that robustness to base-model attacks alone is insufficient. The target model must also exhibit a degree of general adversarial robustness. Because the surrogate model trained by an adaptive attacker provides near-white-box access to our target model, $f_t$ and $f_s$ may still share overlapping vulnerable directions in the perturbation space.


As discussed earlier, Standard-AFT is effective in white-box settings, which represent the worst-case attack scenario. In our setting, however, Standard-AFT is suboptimal because it enforces robustness over the entire perturbation space, namely perturbations generated using the target model’s gradients at each training step. This can overly constrain the model and is unnecessary for our threat model. To address this, we propose \emph{HybridGrad-AFT}, which provides targeted general robustness by restricting the perturbation space to \emph{fundamental and transferable perturbations} rather than the full space. This design aligns with our goal of suppressing transferability: the most critical threats in our threat model are adaptive attacks, whose perturbations must still transfer from a surrogate to the target and therefore lie primarily in this fundamental, transferable subspace.

In HybridGrad-AFT, we augment the training data with adversarial perturbations called \emph{HybridGrad}. These perturbations are generated using a momentum-based PGD attack, in which each PGD step employs a hybrid gradient formed by combining gradients from the base and target models. A scheduler controls this combination: in early PGD steps, updates rely more heavily on the base-model gradient to ensure effectiveness on the base model, and they gradually shift emphasis toward target-model gradient to promote transferability. Thus, HybridGrad perturbations are simultaneously effective on both the base and target models (see Table~\ref{tab:mix_attack} and details in Appendix~\ref{HybridGrad_Detail}).




Overall, HybridGrad-AFT focuses adversarial training on fundamental and transferable perturbations via HybridGrad. As shown in Table~\ref{tab:aft_variants}, it substantially improves robustness against adaptive attacks. Appendix~\ref{HybridGrad-AFT} theoretically formalizes this perturbation space and shows how HybridGrad-AFT focuses training on the fundamental perturbations responsible for transferability across models. We further examine this attack's strength relative to a true white-box adversary, and the effect of surrogate data fidelity, in Appendix~\ref{appendix:whitebox_on_target} and Appendix~\ref{appendix:surrogate_fidelity}, respectively.

\subsection{TransferBreaker}
TransferBreaker is a fine-tuning framework designed to suppress adversarial transferability. It fine-tunes a target model $f_t(\cdot)$, initialized from the base model at the start, and fine-tunes it using a mixture of clean samples, base model adversarial examples, and HybridGrad-augmented adversarial samples, together with LJ-Regularization (see Fig. \ref{fig:main_figure}). The overall training objective is
{
\setlength{\abovedisplayskip}{1pt}
\setlength{\belowdisplayskip}{0pt}
\setlength{\abovedisplayshortskip}{1pt}
\setlength{\belowdisplayshortskip}{0pt}
\begin{equation} \label{EQ_total_loss}
\begin{gathered}
\mathcal{L}_{\text{Total}}
=
\mathcal{L}\big(f_t(x), y\big)
+\lambda_1\, \mathcal{L}\big(f_t(x+\delta), y\big)
+\lambda_2\, \frac{1}{d}\big\| h_t(x+\delta) - h_b(x) \big\|_2^2
\\[0pt]
\text{with}\quad I \sim \text{Bernoulli}(p) \quad \text{and} \quad
\delta = I\,\delta_{\text{hybrid}} + (1-I)\,\delta_{\text{base}} \, .
\end{gathered}
\end{equation}
}

By minimizing $\mathcal{L}_{\text{Total}}$, TransferBreaker suppresses the transferability of adversarial perturbations from the base model to the target model. While the choice of loss $\mathcal{L}$ depends on the base model’s pretraining objective (e.g., cross-entropy for Whisper or CTC for Wav2Vec2), TransferBreaker remains model-agnostic and architecture-independent.
Accordingly, the Bernoulli variable $I$ introduces randomness in selecting the source of the adversarial perturbation (see Appendix~\ref{appendix:BernoulliMixingReason} for comparison with using both perturbations at once), choosing between $\delta_{\text{base}}$ and $\delta_{\text{hybrid}}$ with probability $p$ (set to 0.5 in our experiments). The hyperparameters $\lambda_1$ and $\lambda_2$ control the contributions of the adversarial loss and LJ-Regularization, respectively. Unless otherwise stated, $\lambda_1$ is set to $4$ and $\lambda_2$ is set to $2.5$. We provide an ablation study over these hyperparameters in Appendix \ref{Sec:Ablation_Lambdas}.

\section{Experiments}\label{sec:experiments}




We extensively evaluate TransferBreaker on three typologically diverse languages (Portuguese, Polish, and Arabic) using data from Common Voice 24 (details on the dataset and language selection are provided in Appendix~\ref{appendix_dataset_detail} and Appendix~\ref{Appendix:Language_Selection}). Experiments are conducted across four state-of-the-art ASR backbones: Whisper-Medium, Wav2Vec2-Large-XLSR, MMS-1B, and Whisper-Large-V3-Turbo (model details in Appendix~\ref{app:model_detail}). Table~\ref{tab:main_results} reports results for four model variants: base, fine-tuned (FT), standard adversarially fine-tuned (Std-AFT), and TransferBreaker (Ours). Models are evaluated on both \emph{Clean} data and \emph{Adversarial} data generated on the base model and transferred to the target models. We report WER and CER. Attacks crafted on the base model transfer strongly to FT models, resulting in substantial performance degradation (e.g., for Arabic, the WER of Whisper-Medium increases from $24.7\%$ to $110.0\%$). Although Std-AFT improves robustness, it remains vulnerable. In contrast, TransferBreaker consistently achieves the lowest WER and CER under adversarial evaluation while maintaining competitive clean performance. Its consistent gains across languages (including low-resource Finnish and Vietnamese, Appendix~ \ref{appendix:eval_extremly_low_resource}) and backbones demonstrate that it is both language-agnostic and model-agnostic, making it a broadly applicable defense against transferred adversarial attacks. 

\textbf{Implementation Detail.}
All experiments use the Mozilla Common Voice~24 dataset. Each model is fine-tuned for 4 epochs using AdamW with a learning rate of $1\times10^{-5}$ and a 10\% linear warm-up followed by linear decay. Evaluation is performed using our Advance-PGD attack (Appendix~\ref{appendix:AdvancePGD}). Further implementation details and evaluation metric definitions are provided in Appendices~\ref{Appendix:Extra_Implemetation_Detail} and~\ref{Appendix:Evaluation_Metrics}, respectively. We also provide a computational cost analysis of training in Appendix~\ref{appendix:training_cost_comparison}.


\textbf{Analyzing Results.} 
Averaged across all languages and models, transferred attacks severely degrade fine-tuned models, increasing WER from 16.4 to 92.6 and CER from 4.3 to 48.3. Standard-AFT partially mitigates this, reducing adversarial WER to 42.8 (clean WER 20.3, clean CER 5.7; adversarial CER 16.4). In contrast, TransferBreaker provides stronger robustness, lowering adversarial WER to 27.8 (clean WER 18.9, clean CER 5.2; adversarial CER 9.0), corresponding to a 64.8-point improvement over fine-tuned variants under attack. While both Standard-AFT and TransferBreaker exhibit a trade-off between clean accuracy and robustness \cite{pmlr-v119-raghunathan20a,zhang2019theoretically,tsipras2019robustnessoddsaccuracy}, TransferBreaker achieves significantly higher robustness than Standard-AFT while also maintaining better average clean performance.

\begin{table}[t]
\centering
\begin{minipage}[t]{0.49\textwidth}
\centering
\caption{Ablation of Latent Jacobian Regularization. WER/CER under clean and transferred adversarial settings for Polish (PL), Portuguese (PT), and Arabic (AR) across ASR backbones, with and without LJ-Regularization (LJR). Lower is better.}
\resizebox{1.0\linewidth}{!}{%
\begin{tabular}{
  BBB c
  c >{\columncolor[HTML]{e5e5ff}}c
  c >{\columncolor[HTML]{e5e5ff}}c
  c >{\columncolor[HTML]{e5e5ff}}c
}

\specialrule{1.5pt}{\aboverulesep}{\belowrulesep}

 &  &  &
 & \multicolumn{2}{c}{\textbf{Whisper Medium}}
 & \multicolumn{2}{c}{\textbf{Wav2Vec2 XLSR}}
 & \multicolumn{2}{c}{\textbf{MMS 1B}} \\
\cmidrule(lr){5-6}\cmidrule(lr){7-8}\cmidrule(lr){9-10}
Lang & Setup & Metric
 &
 & \textbf{ w/o LJR } & \textbf{ w/ LJR }
 & \textbf{ w/o LJR } & \textbf{\ \ w/ LJR\ \ }
 & \textbf{ w/o LJR } & \textbf{\ \ w/ LJR\ \ } \\

\specialrule{1.5pt}{\aboverulesep}{\belowrulesep}

\multicolumn{1}{c}{\multirow{4}{*}{\textbf{PL}}}
 & \multirow{2}{*}{\color{CG}Clean} & \color{CG}WER &
 & \color{CG}12.8 & \color{CG}12.0 & \color{CG}16.4 & \color{CG}16.8 & \color{CG}10.5 & \color{CG}10.7 \\
 &  & \color{CG}CER &
 & \color{CG}4.1 & \color{CG}3.8 & \color{CG}4.0 & \color{CG}4.2 & \color{CG}2.6 & \color{CG}2.6 \\

\cmidrule(lr){2-3}\cmidrule(lr){5-6}\cmidrule(lr){7-8}\cmidrule(lr){9-10}

 & \multirow{2}{*}{Base-Adv} & WER &
 & 28.4 & \textbf{24.2} & 26.5 & \textbf{24.8} & 19.2 & \textbf{17.2} \\
 &  & CER &
 & 12.4 & \textbf{9.7} & 7.8 & \textbf{6.9} & 5.5 & \textbf{4.8} \\

\specialrule{1.5pt}{\aboverulesep}{\belowrulesep}

\multicolumn{1}{c}{\multirow{4}{*}{\textbf{PT}}}
 & \multirow{2}{*}{\color{CG}Clean} & \color{CG}WER &
 & \color{CG}15.9 & \color{CG}15.1 & \color{CG}23.3 & \color{CG}23.9 & \color{CG}16.5 & \color{CG}16.4 \\
 &  & \color{CG}CER &
 & \color{CG}5.3 & \color{CG}4.8 & \color{CG}6.6 & \color{CG}6.7 & \color{CG}4.4 & \color{CG}4.3 \\

\cmidrule(lr){2-3}\cmidrule(lr){5-6}\cmidrule(lr){7-8}\cmidrule(lr){9-10}

 & \multirow{2}{*}{Base-Adv} & WER &
 & 26.0 & \textbf{21.7} & 34.5 & \textbf{32.7} & 25.2 & \textbf{23.1} \\
 &  & CER &
 & 10.9 & \textbf{8.2} & 11.5 & \textbf{10.7} & 7.9 & \textbf{7.1} \\

\specialrule{1.5pt}{\aboverulesep}{\belowrulesep}

\multicolumn{1}{c}{\multirow{4}{*}{\textbf{AR}}}
 & \multirow{2}{*}{\color{CG}Clean} & \color{CG}WER &
 & \color{CG}27.0 & \color{CG}26.2 & \color{CG}26.0 & \color{CG}26.4 & \color{CG}28.4 & \color{CG}28.5 \\
 &  & \color{CG}CER &
 & \color{CG}8.7 & \color{CG}8.3 & \color{CG}6.1 & \color{CG}6.2 & \color{CG}6.6 & \color{CG}6.6 \\

\cmidrule(lr){2-3}\cmidrule(lr){5-6}\cmidrule(lr){7-8}\cmidrule(lr){9-10}

 & \multirow{2}{*}{Base-Adv} & WER &
 & 41.2 & \textbf{36.5} & 39.1 & \textbf{36.8} & 42.9 & \textbf{38.9} \\
 &  & CER &
 & 14.9 & \textbf{12.2} & 11.7 & \textbf{10.6} & 12.1 & \textbf{10.7} \\

\specialrule{1.5pt}{\aboverulesep}{\belowrulesep}

\end{tabular}
}
\label{tab:ljr_ablation}

\end{minipage}
\hfill
\begin{minipage}[t]{0.49\textwidth}
\centering
\caption{Ablation of base-model perturbations in TransferBreaker. We report WER/CER under clean and transferred adversarial settings for Polish (PL), Portuguese (PT), and Arabic (AR) across ASR backbones, comparing Standard-AFT with the Base-AFT variant. Lower is better.}
\resizebox{1.0\linewidth}{!}{%
\begin{tabular}{
  BBB c
  c >{\columncolor[HTML]{e5e5ff}}c
  c >{\columncolor[HTML]{e5e5ff}}c
  c >{\columncolor[HTML]{e5e5ff}}c
}

\specialrule{1.5pt}{\aboverulesep}{\belowrulesep}

 &  &  &
 & \multicolumn{2}{c}{\textbf{Whisper Medium}}
 & \multicolumn{2}{c}{\textbf{Wav2Vec2 XLSR}}
 & \multicolumn{2}{c}{\textbf{MMS 1B}} \\
\cmidrule(lr){5-6}\cmidrule(lr){7-8}\cmidrule(lr){9-10}
Lang & Setup & Metric
 &
 & \textbf{Std-AFT} & \textbf{Base-AFT}
 & \textbf{Std-AFT} & \textbf{Base-AFT}
 & \textbf{Std-AFT} & \textbf{Base-AFT} \\

\specialrule{1.5pt}{\aboverulesep}{\belowrulesep}

\multicolumn{1}{c}{\multirow{4}{*}{\textbf{PL}}}
 & \multirow{2}{*}{\color{CG}Clean} & \color{CG}WER &
 & \color{CG}12.9 & \color{CG}11.0 & \color{CG}17.3 & \color{CG}15.9 & \color{CG}12.4 & \color{CG}10.6 \\
 &  & \color{CG}CER &
 & \color{CG}4.2 & \color{CG}3.4 & \color{CG}4.5 & \color{CG}3.9 & \color{CG}3.2 & \color{CG}2.6 \\

\cmidrule(lr){2-3}\cmidrule(lr){5-6}\cmidrule(lr){7-8}\cmidrule(lr){9-10}

 & \multirow{2}{*}{Base-Adv} & WER &
 & 44.0 & \textbf{22.4} & 33.7 & \textbf{24.0} & 28.7 & \textbf{17.2} \\
 &  & CER &
 & 22.5 & \textbf{8.3} & 10.3 & \textbf{6.9} & 8.8 & \textbf{4.9} \\

\specialrule{1.5pt}{\aboverulesep}{\belowrulesep}

\multicolumn{1}{c}{\multirow{4}{*}{\textbf{PT}}}
 & \multirow{2}{*}{\color{CG}Clean} & \color{CG}WER &
 & \color{CG}16.1 & \color{CG}14.2 & \color{CG}26.4 & \color{CG}23.2 & \color{CG}18.2 & \color{CG}16.2 \\
 &  & \color{CG}CER &
 & \color{CG}5.5 & \color{CG}4.4 & \color{CG}7.7 & \color{CG}6.6 & \color{CG}4.9 & \color{CG}4.1 \\

\cmidrule(lr){2-3}\cmidrule(lr){5-6}\cmidrule(lr){7-8}\cmidrule(lr){9-10}

 & \multirow{2}{*}{Base-Adv} & WER &
 & 42.9 & \textbf{19.6} & 41.9 & \textbf{32.7} & 34.5 & \textbf{22.2} \\
 &  & CER &
 & 21.1 & \textbf{6.8} & 14.1 & \textbf{10.6} & 11.5 & \textbf{6.6} \\

\specialrule{1.5pt}{\aboverulesep}{\belowrulesep}

\multicolumn{1}{c}{\multirow{4}{*}{\textbf{AR}}}
 & \multirow{2}{*}{\color{CG}Clean} & \color{CG}WER &
 & \color{CG}28.1 & \color{CG}25.6 & \color{CG}28.2 & \color{CG}25.4 & \color{CG}31.0 & \color{CG}28.2 \\
 &  & \color{CG}CER &
 & \color{CG}8.5 & \color{CG}7.3 & \color{CG}6.8 & \color{CG}5.9 & \color{CG}7.4 & \color{CG}6.5 \\

\cmidrule(lr){2-3}\cmidrule(lr){5-6}\cmidrule(lr){7-8}\cmidrule(lr){9-10}

 & \multirow{2}{*}{Base-Adv} & WER &
 & 58.4 & \textbf{33.4} & 48.5 & \textbf{36.1} & 50.9 & \textbf{38.1} \\
 &  & CER &
 & 27.1 & \textbf{10.3} & 14.3 & \textbf{10.4} & 14.1 & \textbf{10.3} \\

\specialrule{1.5pt}{\aboverulesep}{\belowrulesep}

\end{tabular}
}
\label{tab:base_ablation}

\end{minipage}
\vspace{-10pt}
\end{table}

\section{Ablation Study}



\textbf{Base-AFT.}
To evaluate the role of Base-AFT in TransferBreaker, we conduct an ablation in which the model is fine-tuned on base-model perturbations and clean data. Specifically, in Eq.~\ref{EQ_total_loss} we set $I$ and $\lambda_2$ to zero (Table~\ref{tab:base_ablation}). Across Polish, Portuguese, and Arabic, Base-AFT alone substantially improves robustness to transferred adversarial attacks. Compared to Standard-AFT, it yields consistent gains, demonstrating that this component is critical for suppressing adversarial transferability.

\textbf{Latent Jacobian Regularization.}
To assess the effect of LJ-Regularization, we conduct an ablation by setting $\lambda_2=0$ in Eq.~\ref{EQ_total_loss}, which removes the latent alignment term $\frac{1}{d}\,\| h_t(x+\delta) - h_b(x) \|_2^2$ (Table~\ref{tab:ljr_ablation}). Across Polish, Portuguese, and Arabic, removing this constraint reduces resistance to attacks transferred from the base model, confirming that LJ-Regularization limits adversarial transferability and enhances robustness. In addition, Appendix Table~\ref{tab:jacobian_ljr} shows a decrease in $\mathbb{E}_x\!\left[\|J_{h_t}(x)\delta\|_2^2\right]$, in line with the analysis in Section~\ref{LJR_theory}, which links clean-to-adversarial latent displacement to directional latent-Jacobian sensitivity.


\textbf{HybridGrad-AFT.}
As shown in Table~\ref{tab:aft_variants}, Base-AFT alone provides insufficient robustness against adaptive attacks. Moreover, simply adding Standard-AFT to Base-AFT is suboptimal. A full four-way component ablation in Appendix~\ref{appendix:component_ablation} (Table~\ref{tab:component_ablation}) separately compares Base-AFT, Base-AFT+LJR, Base-AFT+HybridGrad-AFT, and the full framework across Polish and Portuguese. It shows that HybridGrad-AFT substantially improves robustness to adaptive attacks but introduces a trade-off under base-model attacks, while LJR provides complementary gains and partially recovers this loss. Thus, TransferBreaker balances robustness across Base-Adv and Surrogate-Adv rather than uniformly dominating the specialized variants in every setting. (See Appendix~\ref{appendix:TradeOffAccRob} for details.)

\textbf{Additional Ablation Studies.} Further ablation studies are provided in Appendix~\ref{appendix:additional_ablation}. These experiments examine the effectiveness of the proposed HybridGrad perturbations, evaluate robustness under varying signal-to-noise ratio (SNR) constraints, conduct a perceptual evaluation of the adversarial audio, and compare our Advance-PGD attack against strong baselines such as PGD1000, $C\&W$, and AutoAttack. We also analyze the impact of key hyperparameters, including the HybridGrad mixing ratio $\gamma$, the loss coefficients $\lambda_1$ and $\lambda_2$, the effect of removing Bernoulli sampling, and the effect of the HybridGrad scheduler. We additionally evaluate robustness to common non-adversarial acoustic corruptions in Appendix~\ref{appendix:corruption_robustness}. Appendix~\ref{appendix:error_modes} further characterizes the underlying error modes.

\section{Conclusion}


TransferBreaker is a fine-tuning framework for suppressing adversarial transferability from public base models to downstream target models. It combines Base-AFT, HybridGrad-AFT, and Latent Jacobian Regularization to improve robustness to base-model transferred and defense-aware surrogate attacks while largely preserving clean performance. Across multiple ASR backbones and languages, our experiments, ablations, and theoretical analyses clarify the complementary roles of these components and show that explicitly targeting adversarial transferability provides benefits beyond standard adversarial fine-tuning. Overall, these results establish adversarial transferability as an important security consideration when fine-tuning publicly available pretrained ASR models.

\begin{ack}
This work was funded in part by the Swiss National Science Foundation under the ERC-Advanced grant BALM, Swiss National Science Foundation Agreement \#MB25.00024.
\end{ack}


{
\small
\bibliographystyle{unsrtnat}
\bibliography{ref}
}

\newpage
\appendix
\tableofcontents
\newpage
\section{Algorithms and Supporting Jacobian Sensitivity Analysis}
\label{appendix:algorithms_jacobian}

Algorithms~\ref{alg:delta_base} and~\ref{alg:delta_HybridGrad} summarize the procedures used to generate $\delta_{\mathrm{base}}$ and $\delta_{\mathrm{hybrid}}$, respectively.

\begin{algorithm}[H]
\caption{Computing $\delta_{\text{base}}$}
\label{alg:delta_base}
\small
\begin{algorithmic}[1]
\Require Model $f_b$, data $(x,y)$, steps $n$, step size $\alpha$, momentum $\mu$, budget $\epsilon$
\State $\delta_0 \gets 0,\quad m_0 \gets 0$
\For{$j = 0$ to $n-1$}
    \State $g \gets \nabla_{\delta}\mathcal{L}\!\left(f_b(x+\delta_j), y\right)$
    \State $m_{j+1} \gets \mu m_j + \dfrac{g}{\lVert g\rVert_1}$
    \State $\delta_{j+1} \gets \operatorname{clamp}\!\Big(\delta_j + \alpha\,\operatorname{sign}(m_{j+1}),\, -\epsilon,\, \epsilon\Big)$
\EndFor
\State \Return $\delta_{\text{base}} \gets \delta_n$
\end{algorithmic}
\end{algorithm}

\begin{algorithm}[H]
\caption{Computing $\delta_{\mathrm{hybrid}}$}
\label{alg:delta_HybridGrad}
\begin{algorithmic}[1]
\Require Models $f_t$ and $f_b$, data $(x,y)$, steps $n$, step size $\alpha$, momentum $\mu$, HybridGrad ratio $\gamma$, budget $\epsilon$
\State $\delta_0 \gets 0,\quad m_0 \gets 0$
\For{$j=0$ to $n-1$}
    \State $t \gets \dfrac{j+1}{n}$
    \State $\omega_b \gets 1 - \gamma t$
    \State $\omega_t \gets \gamma t$
    \State $g_b \gets \nabla_\delta \mathcal{L}(f_b(x+\delta_j), y)$
    \State $g_t \gets \nabla_\delta \mathcal{L}(f_t(x+\delta_j), y)$
    \State $m_{j+1} \gets \mu m_j + \Big( \omega_b \frac{g_b}{\lVert g_b\rVert_1} + \omega_t \frac{g_t}{\lVert g_t\rVert_1} \Big)$
    \State $\delta_{j+1} \gets \operatorname{clamp}(\delta_j + \alpha\,\operatorname{sign}(m_{j+1}), -\epsilon, \epsilon)$
\EndFor
\State \Return $\delta_{\mathrm{hybrid}} \gets \delta_n$
\end{algorithmic}
\end{algorithm}

Table~\ref{tab:jacobian_ljr} provides complementary evidence for the LJ-Regularization mechanism by reporting the expected directional latent sensitivity $\mathbb{E}_x[\|J_{h_t}(x)\delta\|_2^2]$ with and without LJ-Regularization. The observed reduction is in line with the analysis in Section~\ref{LJR_theory}, which links clean-to-adversarial latent displacement to directional latent-Jacobian sensitivity.

\begin{table}[H]
\centering
\caption{Expected latent sensitivity to input perturbations $\mathbb{E}_x[\|J_{h_t}(x)\delta\|_2^2]$ across ASR backbones, with and without LJ-Regularization (LJR). Lower is better.}
\resizebox{0.7\linewidth}{!}{%
\begin{tabular}{
  BB c
  c >{\columncolor[HTML]{e5e5ff}}c
  c >{\columncolor[HTML]{e5e5ff}}c
  c >{\columncolor[HTML]{e5e5ff}}c
}

\specialrule{1.5pt}{\aboverulesep}{\belowrulesep}

 &  &
 & \multicolumn{2}{c}{\textbf{Whisper Medium}}
 & \multicolumn{2}{c}{\textbf{Wav2Vec2 Large XLSR}}
 & \multicolumn{2}{c}{\textbf{MMS 1B}} \\
\cmidrule(lr){4-5}\cmidrule(lr){6-7}\cmidrule(lr){8-9}
Language & Metric
 &
 & \textbf{w/o LJR} & \textbf{w/ LJR}
 & \textbf{w/o LJR} & \textbf{w/ LJR}
 & \textbf{w/o LJR} & \textbf{w/ LJR} \\

\specialrule{1.5pt}{\aboverulesep}{\belowrulesep}

\multicolumn{1}{c}{\multirow{1}{*}{\textbf{Polish}}}
 & $\mathbb{E}_x\!\left[\|J_{h_t}(x)\delta\|_2^2\right]$ &
 & 438{,}244 & \textbf{147{,}456} & 5{,}041 & \textbf{529} & 30{,}976 & \textbf{17{,}424} \\

\specialrule{1.5pt}{\aboverulesep}{\belowrulesep}

\multicolumn{1}{c}{\multirow{1}{*}{\textbf{Portuguese}}}
 & $\mathbb{E}_x\!\left[\|J_{h_t}(x)\delta\|_2^2\right]$ &
 & 338{,}724 & \textbf{96{,}721} & 3{,}600 & \textbf{625} & 28{,}224 & \textbf{12{,}769} \\

\specialrule{1.5pt}{\aboverulesep}{\belowrulesep}

\end{tabular}
}
\label{tab:jacobian_ljr}
\end{table}










\section{HybridGrad-AFT: Theoretical Insights}\label{HybridGrad-AFT}

HybridGrad-AFT constructs adversarial perturbations by combining gradients from the base and target models. This design is intended to target the shared adversarial subspace, where transferability between the two models occurs. For an input $x$, let $V_b(x)$ and $V_t(x)$ denote the adversarial perturbation sets of the base and target models, respectively. We use $V_F(x)$ to denote the set of \emph{fundamental adversarial perturbations}, whose adversarial effect on the base model persists after fine-tuning and therefore contributes to transferability between the base and target models. By definition, such perturbations are effective on both models, so
\begin{equation}
V_F(x) \subseteq V_b(x) \cap V_t(x) \, .
\label{eq:fundamental-subset}
\end{equation}
Thus, the shared adversarial subspace contains the fundamental perturbations responsible for transferability, while it may also contain vulnerabilities specific to a particular base-target pair.

We next examine how HybridGrad focuses adversarial optimization on this transfer-relevant region. At PGD step $j$, let
\begin{equation}
g_b^{(j)} = \nabla_{\delta}\mathcal{L}\!\left(f_b(x+\delta_j),y\right),
\qquad
g_t^{(j)} = \nabla_{\delta}\mathcal{L}\!\left(f_t(x+\delta_j),y\right),
\end{equation}
and define the normalized directions
\begin{equation}
u_b^{(j)}=\frac{g_b^{(j)}}{\|g_b^{(j)}\|_1},
\qquad
u_t^{(j)}=\frac{g_t^{(j)}}{\|g_t^{(j)}\|_1}.
\end{equation}
As shown in Algorithm~\ref{alg:delta_HybridGrad}, HybridGrad interpolates these directions according to
\begin{equation}
g_H^{(j)}
=
\omega_b^{(j)}u_b^{(j)}
+
\omega_t^{(j)}u_t^{(j)},
\qquad
\omega_t^{(j)}=\gamma\frac{j+1}{n},
\quad
\omega_b^{(j)}=1-\omega_t^{(j)}.
\label{eq:hybridgrad-gradient}
\end{equation}
Early PGD steps therefore emphasize the base-model direction, keeping the perturbation strongly tied to the publicly available base model. As the optimization progresses, the target-model contribution increases, adapting the perturbation toward vulnerabilities that are also present in the target model. For the default $\gamma=0.5$, the final weighting gives equal weight to the two normalized gradients.

This interpolation is directly related to transferability. Under a first-order approximation, for a perturbation $\delta$,
\begin{equation}
\mathcal{L}_b(x+\delta,y)-\mathcal{L}_b(x,y)
\approx
\nabla_x\mathcal{L}_b(x,y)^\top\delta,
\qquad
\mathcal{L}_t(x+\delta,y)-\mathcal{L}_t(x,y)
\approx
\nabla_x\mathcal{L}_t(x,y)^\top\delta.
\label{eq:hybridgrad-firstorder}
\end{equation}
Directions that receive adversarial support from both models can therefore increase the local loss of both the base and target models and are particularly relevant to transferability. By retaining the base-model gradient throughout the attack while progressively incorporating the target-model gradient, HybridGrad favors perturbations that remain effective on both models over perturbations that are specific to only one model. This is consistent with Table~\ref{tab:mix_attack}, where HybridGrad perturbations are effective on both the base and target models.

Accordingly, HybridGrad-AFT focuses adversarial training on a transfer-relevant portion of the perturbation space that contains the fundamental perturbations $V_F(x)$, rather than attempting to cover the entire adversarial space as in Standard-AFT. This provides the theoretical intuition behind HybridGrad-AFT: interpolating the base- and target-model gradients keeps the generated perturbations associated with the base-model threat while adapting them toward vulnerabilities shared with the target model, thereby strengthening robustness along fundamental and transferable adversarial directions.

This analysis characterizes the bias introduced by HybridGrad rather than claiming that every generated perturbation lies exactly in $V_F(x)$ or that HybridGrad recovers the entire fundamental perturbation set.

\section{Adversarial Transferability Across Fine‑Tuning Scales}
\label{appendix:FT_on_more_data}

\begin{figure}[H]
    \centering
    \includegraphics[width=0.7\linewidth]{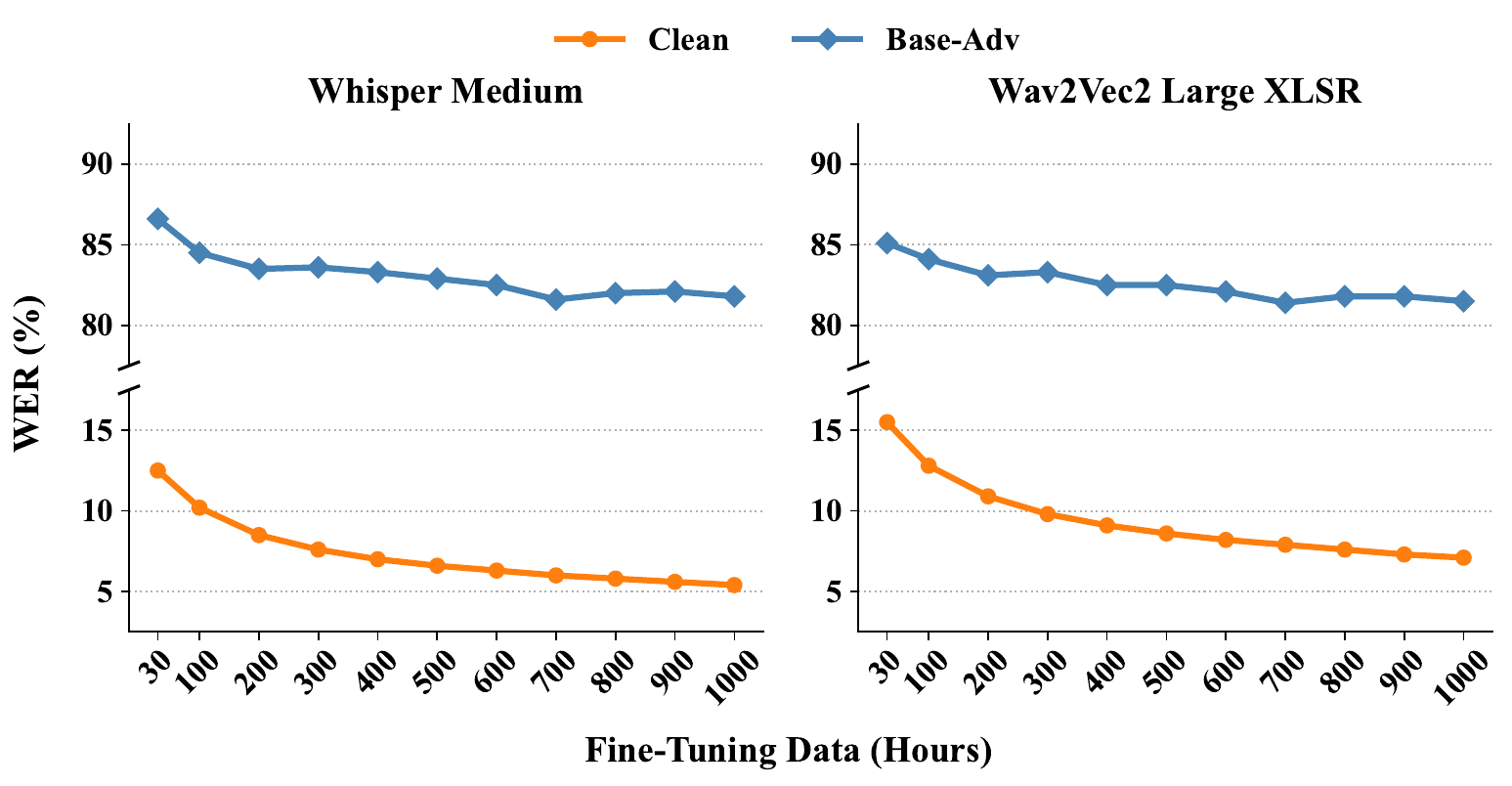}
    \caption{Clean and base-model adversarial attack (Base-Adv) WER across different fine-tuning data sizes from the ParlaSpeech Polish dataset using Whisper-Medium and Wav2Vec2-Large-XLSR. Lower is better.}
    \label{fig:more_FT_fig}
\end{figure}

We fine-tuned Whisper-Medium and Wav2Vec2-Large-XLSR on the Polish subset of the ParlaSpeech dataset \cite{rupnik2022parlaspeech} ($\approx$1{,}010 hours). A 10-hour held-out set was used for evaluation, and all results in Fig.~\ref{fig:more_FT_fig} are reported on this split. The remaining 1{,}000 hours were used for fine-tuning over a single epoch.

To analyze the effect of data scale, we evaluated model checkpoints at 30 hours and then at 100‑hour intervals throughout fine‑tuning and measured WER on clean inputs and under base-model adversarial attacks. As shown in Fig.~\ref{fig:more_FT_fig}, clean WER steadily improves with additional data. In contrast, adversarial WER decreases by only approximately 4--5 percentage points when scaling fine-tuning data from 30 hours to 1{,}000 hours, and remains high ($\approx$82\%).

These results indicate that adversarial transferability from base to fine-tuned models is largely preserved despite substantial increases in fine-tuning data, suggesting that this vulnerability is fundamental rather than a result of insufficient in‑domain data.

\section{Dataset Details}\label{appendix_dataset_detail}

We conduct experiments on three language subsets from Mozilla Common Voice~24: Portuguese, Polish, and Arabic. For each language, we use the predefined \textit{train}, \textit{dev}, and \textit{test} splits provided by the dataset. To ensure consistent experimental conditions across languages, we randomly subsample $\approx$30 hours of utterances from the training split (using 4 epochs, corresponding to $\approx$120 hours of effective training data) and $\approx$4 hours of utterances from the test split for each language.
The total amount of available training data in Common Voice~24 is approximately 26.5 hours for Portuguese (we add a portion of the dev set to bring the total to $\approx$30 hours), 36.3 hours for Polish, and 33.0 hours for Arabic.

In addition, to train surrogate models for adaptive attacks, we use the Multilingual LibriSpeech (MLS) dataset \cite{pratap2020mls} for Portuguese and Polish. Furthermore, the experiment in Section~\ref{appendix:FT_on_more_data} requires a substantially larger amount of training data than is available from Common Voice and MLS, so we additionally use the Polish subset of ParlaSpeech, as described in that section.

\section{Model Details}\label{app:model_detail}
We evaluate three representative ASR model families covering autoregressive and non-autoregressive decoding, as well as spectrogram-based and raw waveform-based input representations. The models differ in architecture, training objectives, and multilingual coverage.

\subsection{Whisper}

Whisper is a multilingual encoder–decoder Transformer ASR model. We use the \textit{medium} and \textit{large-v3 turbo} variants (each roughly 0.8B parameters). The model operates on log-Mel spectrograms and is trained autoregressively with a cross-entropy loss over text tokens. Whisper supports 99 languages. The turbo version is optimized for faster decoding with minimal accuracy degradation.

\subsection{Wav2Vec~2.0}

Wav2Vec~2.0 operates on raw audio waveforms using a convolutional neural network (CNN) feature encoder followed by a Transformer-based encoder. The model performs non-autoregressive decoding and is fine-tuned for ASR using a CTC objective. We use the \textit{large XLSR} variant (approximately 300M parameters), which provides multilingual coverage across 53 languages.

\subsection{MMS}

Massively Multilingual Speech (MMS) is a large-scale multilingual ASR framework supporting more than 1{,}000 languages. The variant used in this work has approximately 1B parameters and consists of an encoder augmented with lightweight language-specific adapters. MMS operates directly on raw audio waveforms and performs non-autoregressive decoding, with  fine-tuning conducted using a CTC objective.

\section{Language Selection Criteria}\label{Appendix:Language_Selection}

We select languages to achieve strong linguistic diversity while maintaining comparable data availability across experimental conditions. Specifically, we choose Portuguese, Polish, and Arabic from Mozilla Common Voice~24 based on three criteria: linguistic diversity, script variation, and balanced data availability. Portuguese is a Romance language, Polish belongs to the Slavic branch of the Indo-European family, and Arabic is a Semitic language from the Afro-Asiatic family. This typological diversity enables evaluation of ASR robustness across distinct language families, reducing the likelihood that observed trends are language-specific. 
The selected languages also differ in writing systems: Portuguese and Polish use the Latin script, while Arabic uses the Arabic script. All three languages have a moderate amount of data within Common Voice~24, avoiding both extremely low- and high-resource conditions, with available training splits ranging from approximately 26.5 to 36.3 hours.

\section{Additional Related Work}\label{Appendix_Additional_Related_Work}
\textbf{Adversarial Attacks on ASR.}
Deep learning–based ASR systems have been shown to be highly vulnerable to adversarial perturbations, which are subtle, often imperceptible modifications to audio inputs that can significantly alter the transcribed output \cite{carlini2018audio, pmlr-v97-qin19a, schoenherr2019adversarial}. Initial approaches primarily targeted models in white-box settings using gradient-based optimization \cite{cisse2017houdini}. More recent work has demonstrated the feasibility of black-box and transfer-based attacks \cite{gao2024transferableadversarialattacksasr,olivier2023transferable,abdullah2021sok}, highlighting a critical reliability issue for ASR systems in sensitive applications such as transcription services, voice assistants, and security systems.

\textbf{Transferability of Adversarial Perturbations in ASR.}
While the transferability of adversarial examples has been extensively studied in computer vision \cite{liu2017delving, gu2024a}, it remains significantly underexplored in automatic speech recognition (ASR). A few prior studies have shown that adversarial perturbations can transfer between different ASR models \cite{gao2024transferableadversarialattacksasr,olivier2023transferable,abdullah2021sok}; however, these evaluations are generally limited in scope and do not consider modern, large-scale ASR systems such as Whisper-large-v3-turbo, an increasingly popular model due to its strong performance across a wide range of languages. Importantly, adversarial transferability in ASR is highly architecture-dependent. In realistic threat scenarios like the one we consider, attackers may possess access to a strong surrogate model that closely resembles the deployed target, necessitating robust defenses tailored to such cases. Despite the widespread practice of fine-tuning public ASR models, prior work has not systematically investigated the transferability of adversarial examples from a pretrained base model to its fine-tuned derivatives. This work fills that gap by demonstrating that adversarial perturbations crafted against a public base model can transfer to private fine-tuned models with near–white-box effectiveness. This finding undermines the assumed protection offered by black-box deployments and exposes a critical vulnerability in real-world ASR pipelines. To address this threat, we propose TransferBreaker, a fine-tuning framework designed to suppress such adversarial transferability.


\textbf{Adversarial Defenses for ASR.}
Defensive strategies against adversarial attacks in automatic speech recognition (ASR) can be broadly grouped into two categories: preprocessing-based defenses \cite{yang2018characterizing, DiffusionDefence} and adversarial training approaches \cite{HybridSpeech,madry2018towards}. Among these, adversarial fine-tuning (AFT), originally proposed in the context of robust optimization for image classification \cite{madry2018towards}, remains the most widely adopted and general-purpose defense. However, preprocessing-based defenses are often bypassed by adaptive attacks and remain fundamentally vulnerable under strong threat models; nevertheless, we also evaluate and compare our method against such defenses in Appendix~\ref{appendix:PreprocessingBaseDefence}. In contrast, AFT offers more principled robustness but presents unique challenges in ASR systems due to the sequential nature of output generation and the high-dimensional prediction space in large models such as Whisper \cite{whisper}. Moreover, standard-AFT assumes full white-box access, which leads to inefficient and overly conservative optimization in settings characterized by transfer-based or black-box threats. In such scenarios, a more targeted defense objective, aligned with the specific threat model, may be significantly more effective and efficient.


\section{Detail of HybridGrad Perturbations}\label{HybridGrad_Detail}
As illustrated in Algorithm~\ref{alg:delta_HybridGrad}, HybridGrad perturbations are generated via an iterative procedure that accumulates a momentum term driven by a \emph{hybrid gradient}. This hybrid gradient is formed by combining gradients computed with respect to a fixed base model and a target model that is being trained. At each iteration, the gradients from both models are normalized and linearly combined, and the resulting direction is integrated into a momentum buffer. This momentum-based update stabilizes the optimization process and promotes consistent adversarial directions across iterations.

A time-dependent scheduling mechanism controls the relative contribution of the two model gradients. By default, the HybridGrad ratio, which governs the contribution of the target-model gradient, is set to \( \gamma = 0.5 \) (see Appendix~\ref{subsec:gamma_ablation}). In the early PGD steps, the update direction is dominated by the base-model gradient, ensuring that the perturbation remains adversarial with respect to the base model. As the optimization progresses, the scheduler gradually increases the influence of the target-model gradient, allowing the perturbation to adapt to the evolving decision boundary of the target model. This gradual transition enables HybridGrad perturbations to maintain effectiveness on the base model while simultaneously aligning with the target model, thereby producing transferable adversarial examples.

\section{Additional Ablation Studies}\label{appendix:additional_ablation}

\subsection{Four-Way Component Ablation}\label{appendix:component_ablation}

Table~\ref{tab:component_ablation} isolates the contributions of Base-AFT, LJR, and HybridGrad-AFT by comparing Base-AFT, Base-AFT+LJR, Base-AFT+HybridGrad-AFT, and the full framework on Polish and Portuguese. Across the two languages, HybridGrad-AFT provides the dominant improvement under Surrogate-Adv, while introducing a moderate trade-off under Base-Adv. LJR provides a complementary effect: added to Base-AFT it strengthens robustness to base-model attacks, and added on top of HybridGrad-AFT it improves the balance between Base-Adv and Surrogate-Adv. These results support the interpretation that Base-AFT primarily targets attacks transferred from the public base model, HybridGrad-AFT targets directions that remain transferable to defense-aware surrogates, and LJR reinforces robustness along the perturbation directions used during fine-tuning.

\begin{table}[h]
\centering
\caption{Four-way component ablation on Polish and Portuguese across three ASR backbones. Full denotes the complete TransferBreaker framework, combining Base-AFT, LJR, and HybridGrad-AFT; Base-Adv and Surrogate-Adv denote attacks generated on the public base model and a defense-aware surrogate, respectively. Lower is better.}
\label{tab:component_ablation}
\resizebox{1.0\linewidth}{!}{%
\begin{tabular}{
  BBB c
  c c c >{\columncolor[HTML]{e5e5ff}}c
  c c c >{\columncolor[HTML]{e5e5ff}}c
  c c c >{\columncolor[HTML]{e5e5ff}}c
}
\specialrule{1.5pt}{\aboverulesep}{\belowrulesep}
 &  &  &
 & \multicolumn{4}{c}{\textbf{Whisper Medium}}
 & \multicolumn{4}{c}{\textbf{Wav2Vec2 Large XLSR}}
 & \multicolumn{4}{c}{\textbf{MMS 1B}} \\
\cmidrule(lr){5-8}\cmidrule(lr){9-12}\cmidrule(lr){13-16}
Language & Setup & Metric &
 & \textbf{Base-AFT} & \makecell{\textbf{Base-AFT} \\ \textbf{+LJR}} & \makecell{\textbf{Base-AFT} \\ \textbf{+HybridGrad-AFT}} & \textbf{Full}
 & \textbf{Base-AFT} & \makecell{\textbf{Base-AFT} \\ \textbf{+LJR}} & \makecell{\textbf{Base-AFT} \\ \textbf{+HybridGrad-AFT}} & \textbf{Full}
 & \textbf{Base-AFT} & \makecell{\textbf{Base-AFT} \\ \textbf{+LJR}} & \makecell{\textbf{Base-AFT} \\ \textbf{+HybridGrad-AFT}} & \textbf{Full} \\
\specialrule{1.5pt}{\aboverulesep}{\belowrulesep}

\multicolumn{1}{c}{\multirow{6}{*}{\textbf{Polish}}}
 & \multirow{2}{*}{\color{CG}Clean} & \color{CG}WER &
 & \color{CG}11.0 & \color{CG}10.8 & \color{CG}12.8 & \color{CG}12.0
 & \color{CG}15.9 & \color{CG}16.1 & \color{CG}16.4 & \color{CG}16.8
 & \color{CG}10.6 & \color{CG}10.6 & \color{CG}10.5 & \color{CG}10.7 \\

 &  & \color{CG}CER &
 & \color{CG}3.4 & \color{CG}3.3 & \color{CG}4.1 & \color{CG}3.8
 & \color{CG}3.9 & \color{CG}3.9 & \color{CG}4.0 & \color{CG}4.2
 & \color{CG}2.6 & \color{CG}2.7 & \color{CG}2.6 & \color{CG}2.6 \\

\cmidrule(lr){2-3}\cmidrule(lr){5-8}\cmidrule(lr){9-12}\cmidrule(lr){13-16}

 & \multirow{2}{*}{Base-Adv} & WER &
 & 22.4 & 19.5 & 28.4 & 24.2
 & 24.0 & 22.5 & 26.5 & 24.8
 & 17.2 & 15.6 & 19.2 & 17.2 \\

 &  & CER &
 & 8.3 & 7.0 & 12.4 & 9.7
 & 6.9 & 6.4 & 7.8 & 6.9
 & 4.9 & 4.4 & 5.5 & 4.8 \\

\cmidrule(lr){2-3}\cmidrule(lr){5-8}\cmidrule(lr){9-12}\cmidrule(lr){13-16}

 & \multirow{2}{*}{Surrogate-Adv} & WER &
 & 57.6 & 56.6 & 45.8 & 44.1
 & 54.2 & 53.2 & 42.5 & 40.7
 & 48.6 & 47.9 & 38.5 & 37.2 \\

 &  & CER &
 & 29.6 & 28.8 & 21.6 & 20.5
 & 23.5 & 22.8 & 16.3 & 15.3
 & 19.7 & 19.3 & 14.2 & 13.3 \\

\specialrule{1.5pt}{\aboverulesep}{\belowrulesep}

\multicolumn{1}{c}{\multirow{6}{*}{\textbf{Portuguese}}}
 & \multirow{2}{*}{\color{CG}Clean} & \color{CG}WER &
 & \color{CG}14.2 & \color{CG}14.2 & \color{CG}15.9 & \color{CG}15.1
 & \color{CG}23.2 & \color{CG}23.4 & \color{CG}23.3 & \color{CG}23.9
 & \color{CG}16.2 & \color{CG}16.1 & \color{CG}16.5 & \color{CG}16.4 \\

 &  & \color{CG}CER &
 & \color{CG}4.4 & \color{CG}4.3 & \color{CG}5.3 & \color{CG}4.8
 & \color{CG}6.6 & \color{CG}6.6 & \color{CG}6.6 & \color{CG}6.7
 & \color{CG}4.1 & \color{CG}4.1 & \color{CG}4.4 & \color{CG}4.3 \\

\cmidrule(lr){2-3}\cmidrule(lr){5-8}\cmidrule(lr){9-12}\cmidrule(lr){13-16}

 & \multirow{2}{*}{Base-Adv} & WER &
 & 19.6 & 17.9 & 26.0 & 21.7
 & 32.7 & 31.1 & 34.5 & 32.7
 & 22.2 & 20.6 & 25.2 & 23.1 \\

 &  & CER &
 & 6.8 & 5.9 & 10.9 & 8.2
 & 10.6 & 10.2 & 11.5 & 10.7
 & 6.6 & 6.2 & 7.9 & 7.1 \\

\cmidrule(lr){2-3}\cmidrule(lr){5-8}\cmidrule(lr){9-12}\cmidrule(lr){13-16}

 & \multirow{2}{*}{Surrogate-Adv} & WER &
 & 56.4 & 56.2 & 45.4 & 44.7
 & 62.8 & 62.5 & 50.3 & 48.9
 & 60.3 & 59.3 & 44.3 & 42.8 \\

 &  & CER &
 & 27.8 & 27.5 & 21.2 & 20.8
 & 27.7 & 27.4 & 18.8 & 18.1
 & 27.6 & 26.9 & 18.0 & 17.0 \\

\specialrule{1.5pt}{\aboverulesep}{\belowrulesep}
\end{tabular}
}
\end{table}

\subsection{Frozen-Base Anchor versus Stop-Gradient Self-Invariance}\label{appendix:ljr_anchor_ablation}

Section~\ref{LJR_theory} shows that the frozen-base outer update consists of a stop-gradient self-invariance update together with a clean-representation drift correction anchored to the frozen base model. To evaluate the effect of this frozen-base reference, we compare three variants: removing LJR (w/o), using the stop-gradient self-invariance objective in Eq.~\ref{eq:self_ljr} (Self), and using the frozen-base objective in Eq.~\ref{LJR_main_formula} (Base). In Self, the stop-gradient operator prevents the clean branch from being optimized through the regularizer, avoiding the trivial representation-collapse solution that can arise when a squared-distance objective backpropagates through both branches \cite{grill2020bootstrap,chen2021exploring}. All other training settings are unchanged.

Table~\ref{tab:ljr_anchor_ablation} shows that Self improves robustness to transferred attacks relative to removing LJR, while Base further reduces adversarial WER/CER for both Whisper-Medium and Wav2Vec2 on Polish and Portuguese. Clean performance remains comparable across the three variants. Since Self and Base differ only in the reference used by the regularizer, this comparison isolates the effect of using the frozen base representation instead of the target model's own clean representation. The consistent improvement in adversarial WER/CER supports our use of the frozen base model as the reference for the base-to-target transfer setting.

\begin{table}[H]
\centering
\caption{Frozen-base anchoring versus stop-gradient self-invariance. ``w/o'' removes LJR, ``Self'' uses the stop-gradient objective in Eq.~\ref{eq:self_ljr}, and ``Base'' uses the frozen-base objective in Eq.~\ref{LJR_main_formula}. Values are WER/CER; lower is better.}
\resizebox{0.75\linewidth}{!}{%
\begin{tabular}{ll cc>{\columncolor[HTML]{e5e5ff}}c cc>{\columncolor[HTML]{e5e5ff}}c}
\specialrule{1.5pt}{\aboverulesep}{\belowrulesep}
& & \multicolumn{3}{c}{\textbf{Whisper Medium}} & \multicolumn{3}{c}{\textbf{Wav2Vec2 Large XLSR}} \\
\cmidrule(lr){3-5}\cmidrule(lr){6-8}
\textbf{Language} & \textbf{Setup} & \textbf{w/o} & \textbf{Self} & \textbf{Base (ours)} & \textbf{w/o} & \textbf{Self} & \textbf{Base (ours)} \\
\specialrule{1.5pt}{\aboverulesep}{\belowrulesep}
\multirow{2}{*}{\textbf{Polish}}
& \textcolor{gray}{Clean} & \textcolor{gray}{12.8/4.1} & \textcolor{gray}{12.5/3.9} & \textcolor{gray}{12.0/3.8} & \textcolor{gray}{16.4/4.0} & \textcolor{gray}{16.6/4.1} & \textcolor{gray}{16.8/4.2} \\
& Base-Adv & 28.4/12.4 & 26.6/11.2 & \textbf{24.2/9.7} & 26.5/7.8 & 25.9/7.5 & \textbf{24.8/6.9} \\
\specialrule{1.0pt}{\aboverulesep}{\belowrulesep}
\multirow{2}{*}{\textbf{Portuguese}}
& \textcolor{gray}{Clean} & \textcolor{gray}{15.9/5.3} & \textcolor{gray}{15.5/5.0} & \textcolor{gray}{15.1/4.8} & \textcolor{gray}{23.3/6.6} & \textcolor{gray}{23.6/6.7} & \textcolor{gray}{23.9/6.7} \\
& Base-Adv & 26.0/10.9 & 24.0/9.5 & \textbf{21.7/8.2} & 34.5/11.5 & 33.8/11.2 & \textbf{32.7/10.7} \\
\specialrule{1.5pt}{\aboverulesep}{\belowrulesep}
\end{tabular}%
}
\label{tab:ljr_anchor_ablation}
\end{table}

\subsection{Evaluation Under Different Signal-to-Noise Ratios (SNR)}\label{subsec:snr_eval}
\begin{figure}[H]
    \centering
    \includegraphics[width=0.7\linewidth]{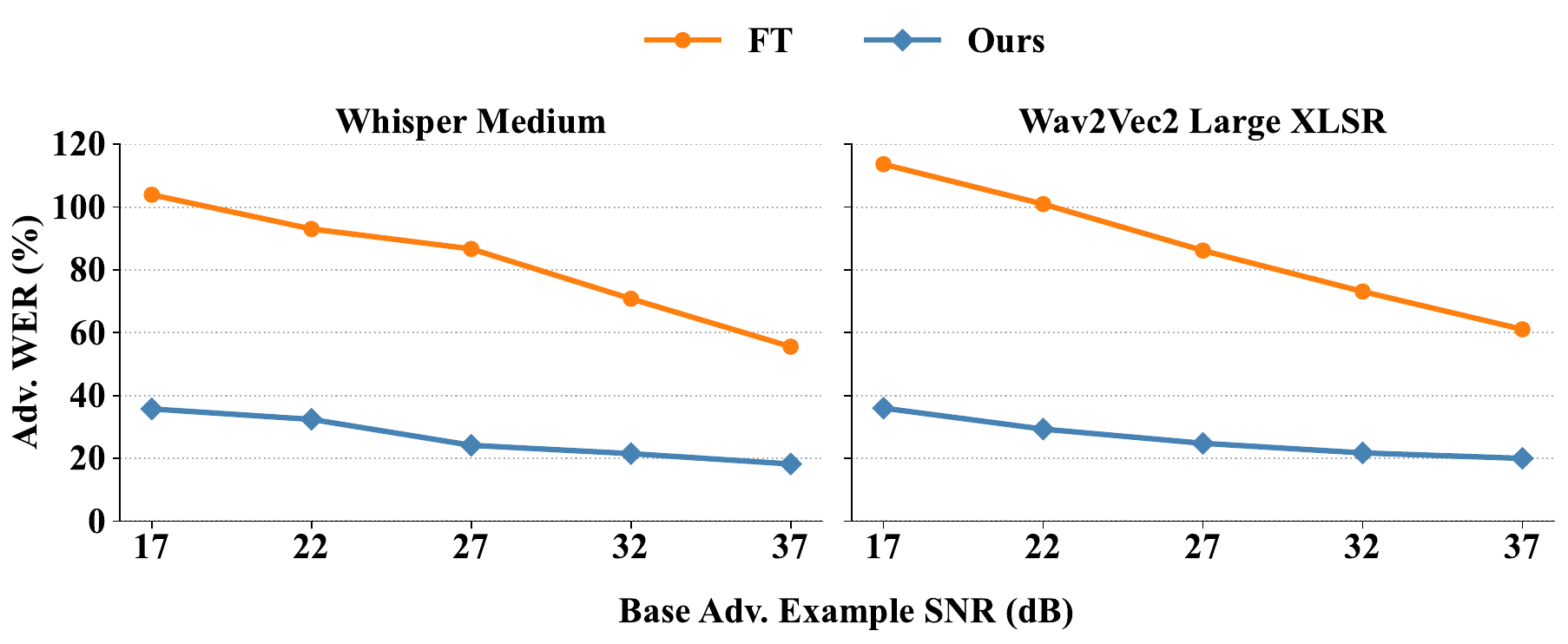}
    \caption{Performance under base‑model adversarial attacks across varying SNR levels, evaluated on Polish using Whisper‑Medium and Wav2Vec2‑Large‑XLSR. Lower is better.}
    \label{fig:snr_figure}
\end{figure}

To assess the robustness of ASR models under varying perturbation strengths, we evaluate adversarial examples generated under different signal-to-noise ratio (SNR) constraints. The SNR directly controls the perceptibility of adversarial noise: lower SNR values correspond to stronger, more perceptible perturbations, whereas higher SNR values yield weaker and less noticeable noise.
We consider five SNR thresholds: 17, 22, 27, 32, and 37~dB. For each threshold, adversarial examples are generated using the base model and subsequently evaluated on both the fine-tuned (FT) and TransferBreaker (Ours) variants. Figure~\ref{fig:snr_figure} presents the results for Polish using Whisper-Medium and Wav2Vec2-Large-XLSR.
    
\subsection{Additional Baseline Attack Adaptations and Comparison}\label{subsec:other_attacks}
To further validate the effectiveness of the proposed defense and the attack used for evaluation, we compare against several strong adversarial attack baselines. Specifically, we evaluate our proposed A-PGD attack alongside three widely used and competitive methods: PGD1000, Carlini \& Wagner (C\&W) \cite{carlini2017towards}, and AutoAttack \cite{croce2020reliable}. Since C\&W and AutoAttack were originally formulated for image classification, we detail below how each is adapted to the sequence-based, variable-length setting of ASR. All baselines are evaluated under the same per-utterance 27~dB SNR $\varepsilon$-ball, $L_\infty$ projection, and sequence task loss (CTC or teacher-forced cross-entropy) as A-PGD, ensuring that Table~\ref{tab:attack_compare} compares all attacks at equal perturbation energy.

\textbf{PGD1000.} A projected gradient descent attack with 1000 update steps under the fixed perturbation constraint, designed to maximize attack success via standard sign-gradient ascent on the task loss.

\textbf{C\&W.} The standard C\&W logit-margin loss is undefined for variable-length transcriptions, so, following Carlini \& Wagner's audio attack~\cite{carlini2018audio}, we substitute the sequence task loss. Since our attacks are untargeted, we minimize $\lVert \delta \rVert_{2}^{2} - c \cdot \mathcal{L}_{\text{task}}(x+\delta, y)$, i.e., maximal task loss at minimal distortion. Optimization follows the original C\&W recipe: a $\tanh$ change of variables, Adam with learning rate $10^{-3}$, 1000 steps, and a 5-round binary search over $c$, projecting the best-WER iterate onto the $\varepsilon$-ball.

\textbf{AutoAttack.} Two of AutoAttack's four components do not transfer to ASR: the DLR-based targeted step requires an ordering of class logits, and FAB requires a decision boundary, neither of which is well-defined for sequence outputs. We therefore construct an ensemble of the two components that do apply: (i) parameter-free PGD with adaptive step size on the task loss (momentum 0.75, step-size halving on loss stagnation, restart from the best iterate), and (ii) Square Attack, adapted following AudioMarkBench~\cite{audiomarkbench} by sampling squares in the time-frequency plane, mapping them back to the waveform via inverse STFT, and projecting onto $\varepsilon$, with WER as the attack objective. We report the per-utterance worst case across both components.

To ensure a fair and rigorous comparison, we carefully tune the hyperparameters of all baseline attacks that require tuning to approach their strongest achievable performance. As shown in Table~\ref{tab:attack_compare}, A-PGD yields higher WER and CER than PGD1000, C\&W, and AutoAttack in most settings, indicating that it constitutes a particularly strong baseline attack for evaluation.
\begin{table}[t]
\centering
\caption{WER/CER (\%) of TransferBreaker (\textbf{Ours}) under transferred attacks (PGD1000, C\&W, AutoAttack, SAGO, M-DI$^2$, ILAP, APGD+DI, and A-PGD) on Polish and Portuguese across ASR backbones. Higher is better (stronger attack against our defense).}\resizebox{0.7\linewidth}{!}{%
\begin{tabular}{BBB c
  c
  c
  c
}
\specialrule{1.5pt}{\aboverulesep}{\belowrulesep}
 &  &  &
 & \textbf{Whisper Medium}
 & \textbf{Wav2Vec2 Large XLSR}
 & \textbf{MMS 1B} \\
\cmidrule(lr){5-5}\cmidrule(lr){6-6}\cmidrule(lr){7-7}
Language & Attack & Metric
 &
 & \textbf{Ours}
 & \textbf{Ours}
 & \textbf{Ours} \\
\specialrule{1.5pt}{\aboverulesep}{\belowrulesep}
\multicolumn{1}{c}{\multirow{16}{*}{\textbf{Polish}}}
 & \multirow{2}{*}{PGD1000} & WER &  & 22.5 & 23.8 & 17.0 \\
 &  & CER &  & 9.0 & 6.6 & 4.7 \\
\cmidrule(lr){2-3}\cmidrule(lr){5-5}\cmidrule(lr){6-6}\cmidrule(lr){7-7}
 & \multirow{2}{*}{C\&W} & WER &  & 23.5 & 23.2 & 16.5 \\
 &  & CER &  & 9.4 & 6.6 & 4.6 \\
\cmidrule(lr){2-3}\cmidrule(lr){5-5}\cmidrule(lr){6-6}\cmidrule(lr){7-7}
 & \multirow{2}{*}{AutoAttack} & WER &  & 22.7 & 22.8 & 16.3 \\
 &  & CER &  & 8.9 & 6.4 & 4.4 \\
\cmidrule(lr){2-3}\cmidrule(lr){5-5}\cmidrule(lr){6-6}\cmidrule(lr){7-7}
 & \multirow{2}{*}{SAGO} & WER &  & 25.0 & 25.4 & 17.8 \\
 &  & CER &  & 10.1 & 7.1 & 5.0 \\
\cmidrule(lr){2-3}\cmidrule(lr){5-5}\cmidrule(lr){6-6}\cmidrule(lr){7-7}
 & \multirow{2}{*}{M-DI$^2$} & WER &  & 25.3 & 25.7 & 18.0 \\
 &  & CER &  & 10.2 & 7.2 & 5.1 \\
\cmidrule(lr){2-3}\cmidrule(lr){5-5}\cmidrule(lr){6-6}\cmidrule(lr){7-7}
 & \multirow{2}{*}{ILAP} & WER &  & 23.6 & 24.1 & 16.9 \\
 &  & CER &  & 9.3 & 6.7 & 4.6 \\
\cmidrule(lr){2-3}\cmidrule(lr){5-5}\cmidrule(lr){6-6}\cmidrule(lr){7-7}
 & \multirow{2}{*}{APGD+DI} & WER &  & 26.0 & 26.3 & 18.6 \\
 &  & CER &  & 10.6 & 7.5 & 5.3 \\
\cmidrule(lr){2-3}\cmidrule(lr){5-5}\cmidrule(lr){6-6}\cmidrule(lr){7-7}
\rowcolor[HTML]{e5e5ff} &  & WER &  & 24.2 & 24.8 & 17.2 \\
\rowcolor[HTML]{e5e5ff} & \multirow{-2}{*}{A-PGD (Ours)} & CER &  & 9.7 & 6.9 & 4.8 \\
\specialrule{1.5pt}{\aboverulesep}{\belowrulesep}
\multicolumn{1}{c}{\multirow{16}{*}{\textbf{Portuguese}}}
 & \multirow{2}{*}{PGD1000} & WER &  & 25.3 & 32.6 & 23.5 \\
 &  & CER &  & 11.8 & 10.2 & 7.3 \\
\cmidrule(lr){2-3}\cmidrule(lr){5-5}\cmidrule(lr){6-6}\cmidrule(lr){7-7}
 & \multirow{2}{*}{C\&W} & WER &  & 23.7 & 29.4 & 22.9 \\
 &  & CER &  & 10.2 & 8.8 & 7.1 \\
\cmidrule(lr){2-3}\cmidrule(lr){5-5}\cmidrule(lr){6-6}\cmidrule(lr){7-7}
 & \multirow{2}{*}{AutoAttack} & WER &  & 21.4 & 28.7 & 22.5 \\
 &  & CER &  & 7.9 & 8.6 & 6.8 \\
\cmidrule(lr){2-3}\cmidrule(lr){5-5}\cmidrule(lr){6-6}\cmidrule(lr){7-7}
 & \multirow{2}{*}{SAGO} & WER &  & 23.6 & 33.1 & 24.9 \\
 &  & CER &  & 10.1 & 10.7 & 7.4 \\
\cmidrule(lr){2-3}\cmidrule(lr){5-5}\cmidrule(lr){6-6}\cmidrule(lr){7-7}
 & \multirow{2}{*}{M-DI$^2$} & WER &  & 24.0 & 33.3 & 24.0 \\
 &  & CER &  & 10.3 & 11.0 & 7.8 \\
\cmidrule(lr){2-3}\cmidrule(lr){5-5}\cmidrule(lr){6-6}\cmidrule(lr){7-7}
 & \multirow{2}{*}{ILAP} & WER &  & 22.1 & 31.2 & 22.6 \\
 &  & CER &  & 9.2 & 10.6 & 7.0 \\
\cmidrule(lr){2-3}\cmidrule(lr){5-5}\cmidrule(lr){6-6}\cmidrule(lr){7-7}
 & \multirow{2}{*}{APGD+DI} & WER &  & 24.9 & 34.3 & 25.7 \\
 &  & CER &  & 10.5 & 11.3 & 8.1 \\
\cmidrule(lr){2-3}\cmidrule(lr){5-5}\cmidrule(lr){6-6}\cmidrule(lr){7-7}
\rowcolor[HTML]{e5e5ff} &  & WER &  & 21.7 & 32.7 & 23.1 \\
\rowcolor[HTML]{e5e5ff} & \multirow{-2}{*}{A-PGD (Ours)} & CER &  & 8.2 & 10.7 & 7.1 \\
\specialrule{1.5pt}{\aboverulesep}{\belowrulesep}
\end{tabular}
}
\label{tab:attack_compare}
\end{table}

\textbf{Transferability-Boosting Attacks.} Since the choice of attack method is critical to transferability, we further evaluate attacks explicitly designed to enhance cross-model transfer, beyond the iterative-momentum optimization already present in A-PGD. All attacks below use the same 27~dB SNR budget and 50 total iterations as A-PGD, and results against TransferBreaker are reported alongside PGD1000, C\&W, and AutoAttack in Table~\ref{tab:attack_compare}.

\textbf{M-DI\textsuperscript{2}-FGSM and APGD+DI.} We adapt M-DI\textsuperscript{2}-FGSM \cite{xie2019improving}, which applies a random differentiable input transformation at each step, and construct APGD+DI, our A-PGD augmented with the same per-step input transformation. At each step, we randomly apply one mild differentiable transformation, including temporal shifting, random padding and cropping, amplitude scaling, slight resampling, time stretching or pitch shifting, low-level Gaussian noise, and frequency filtering.

\textbf{SAGO.} An ASR-specific attack \cite{gao2024transferableadversarialattacksasr} that uses a voice-activity mask to concentrate perturbations on speech regions.

\textbf{ILAP.} This attack \cite{Huang2019EnhancingAE} amplifies the clean-to-adversarial displacement at an intermediate feature layer while preserving its original adversarial direction. We allocate 25 iterations to generating the initial adversarial example and 25 iterations to intermediate-level refinement, applied to the encoder output in Whisper and to the encoder states immediately before the CTC head in Wav2Vec2 and MMS.

\textbf{MI-CWA (Adaptive Threat Model).} For our adaptive threat model (Section~\ref{sec:adaptive_thread_model}), we additionally evaluate MI-CWA~\cite{chen2024rethinking}, an ensemble-based transfer attack that searches for vulnerabilities shared across multiple surrogate models. CWA combines a sharpness-aware step favoring flatter adversarial regions with a cosine-similarity objective aligning gradients across the surrogate ensemble, while MI-CWA further stabilizes these directions using momentum. We adapt MI-CWA to ASR using CTC or sequence cross-entropy loss and the same 27~dB SNR constraint. The ensemble comprises six checkpoints from different epochs of the adaptive attacker's surrogate-training process, evaluated over 50 outer iterations for a deliberately strong evaluation, requiring approximately 12$\times$ more model-gradient computation than a 50-step single-surrogate attack. Results are reported in Table~\ref{tab:adaptive_attacks}, evaluated on Polish.

\textbf{Takeaway.} Against TransferBreaker, the transferability-boosting attacks (SAGO, M-DI\textsuperscript{2}, ILAP, APGD+DI) yield at most a modest gain over A-PGD (Table~\ref{tab:attack_compare}), and the ensemble-based adaptive attack MI-CWA improves by only +2--3 WER over the single-surrogate adaptive A-PGD while remaining well below the degradation observed against Base-AFT (Table~\ref{tab:adaptive_attacks}). These results indicate that our evaluation attack is not a conveniently weak choice, and that our robustness advantage over Base-AFT persists under stronger and transferability-optimized attacks.



\begin{table}[H]
\centering
\caption{WER/CER (\%) under the ensemble-based adaptive attack MI-CWA, compared to the single-surrogate adaptive A-PGD, on Polish. Lower is better.}
\resizebox{0.7\linewidth}{!}{%
\begin{tabular}{B ccc}
\specialrule{1.5pt}{\aboverulesep}{\belowrulesep}
Model & MI-CWA (Ours) & MI-CWA (Base-AFT) & Surr.-A-PGD (Ours) \\
\specialrule{1.5pt}{\aboverulesep}{\belowrulesep}
Whisper Medium & 46.4/21.9 & 60.2/30.8 & 44.1/20.5 \\
Wav2Vec2 Large XLSR & 42.8/16.3 & 56.8/24.8 & 40.7/15.3 \\
MMS 1B & 39.1/14.2 & 51.4/21.0 & 37.2/13.3 \\
\specialrule{1.5pt}{\aboverulesep}{\belowrulesep}
\end{tabular}
}
\label{tab:adaptive_attacks}
\end{table}

\subsection{Ablation Study on Loss Coefficients}
\label{Sec:Ablation_Lambdas}

This subsection examines the sensitivity of TransferBreaker to its two loss coefficients, $\lambda_1$ and $\lambda_2$, which weight the adversarial loss and the LJ-Regularization term in Eq.~\eqref{EQ_total_loss}, respectively.

\paragraph{Effect of $\lambda_1$.}
We first study the effect of $\lambda_1$, which controls the contribution of adversarial examples during fine-tuning. We evaluate $\lambda_1 \in \{2, 4, 6\}$ while fixing $\lambda_2$ to its default value. Results for Whisper-Medium and Wav2Vec2-Large-XLSR are reported in Table~\ref{tab:ablation_lambda1}. Overall, $\lambda_1 = 4$ achieves the most balanced performance across evaluation settings and is therefore selected as the default value.
\begin{table}[ht]
\centering
\caption{Effect of the adversarial loss coefficient $\lambda_1$ on clean accuracy and adversarial transferability. Results are reported for Polish using Whisper-Medium and Wav2Vec2-Large-XLSR backbones. Lower is better.}
\resizebox{0.7\linewidth}{!}{%
\begin{tabular}{BBB c
  c c c
  c c c
}

\specialrule{1.5pt}{\aboverulesep}{\belowrulesep}

 &  &  &
 & \multicolumn{3}{c}{\textbf{Whisper Medium (Ours)}}
 & \multicolumn{3}{c}{\textbf{Wav2Vec2 Large XLSR (Ours)}} \\
\cmidrule(lr){5-7}\cmidrule(lr){8-10}

Language & Setup & Metric
 &
 & $\lambda_1 = 2$ & $\lambda_1 = 4$ & $\lambda_1 = 6$
 & $\lambda_1 = 2$ & $\lambda_1 = 4$ & $\lambda_1 = 6$ \\

\specialrule{1.5pt}{\aboverulesep}{\belowrulesep}

\multicolumn{1}{c}{\multirow{4}{*}{\textbf{Polish}}}

 & \multirow{2}{*}{Clean} & WER
 &  & 11.5 & 12.0 & 13.2
 & 16.1 & 16.8 & 17.4 \\

 &  & CER
 &  & 3.6 & 3.8 & 4.3
 & 4.0 & 4.2 & 4.4 \\

\cmidrule(lr){2-3}\cmidrule(lr){5-7}\cmidrule(lr){8-10}

 & \multirow{2}{*}{Base-Adv} & WER
 &  & 25.9 & 24.2 & 24.8
 & 25.6 & 24.8 & 24.9 \\

 &  & CER
 &  & 10.5 & 9.7 & 10
 & 7.2 & 6.9 & 6.9 \\

\specialrule{1.5pt}{\aboverulesep}{\belowrulesep}

\end{tabular}
}
\label{tab:ablation_lambda1}
\end{table}

\paragraph{Effect of $\lambda_2$.}
We next analyze the impact of $\lambda_2$, which governs the strength of LJ-Regularization. We vary $\lambda_2 \in \{1, 2.5, 5\}$ while keeping $\lambda_1$ fixed to its default value. The corresponding results are summarized in Table~\ref{tab:ablation_lambda2}. Overall, $\lambda_2 = 2.5$ provides the best trade-off across evaluation settings and is adopted as the default value.
\begin{table}[ht]
\centering
\caption{Effect of the LJ-Regularization coefficient $\lambda_2$ on clean accuracy and adversarial transferability. Results are reported for Polish using Whisper-Medium and Wav2Vec2-Large-XLSR backbones. Lower is better.}
\resizebox{0.7\linewidth}{!}{%
\begin{tabular}{BBB c
  c c c
  c c c
}

\specialrule{1.5pt}{\aboverulesep}{\belowrulesep}

 &  &  &
 & \multicolumn{3}{c}{\textbf{Whisper Medium (Ours)}}
 & \multicolumn{3}{c}{\textbf{Wav2Vec2 Large XLSR (Ours)}} \\
\cmidrule(lr){5-7}\cmidrule(lr){8-10}

Language & Setup & Metric
 &
 & $\lambda_2 = 1$ & $\lambda_2 = 2.5$ & $\lambda_2 = 5$
 & $\lambda_2 = 1$ & $\lambda_2 = 2.5$ & $\lambda_2 = 5$ \\

\specialrule{1.5pt}{\aboverulesep}{\belowrulesep}

\multicolumn{1}{c}{\multirow{4}{*}{\textbf{Polish}}}
 & \multirow{2}{*}{Clean} & WER
 &  & 12.4 & 12.0 & 12.7
 & 16.5 & 16.8 & 17.0 \\

 &  & CER
 &  & 3.9 & 3.8 & 4.0
 & 4.1 & 4.2 & 4.3 \\

\cmidrule(lr){2-3}\cmidrule(lr){5-7}\cmidrule(lr){8-10}

 & \multirow{2}{*}{Base-Adv} & WER
 &  & 25.5 & 24.2 & 24.0
 & 25.2 & 24.8 & 24.7 \\

 &  & CER
 &  & 10.3 & 9.7 & 9.7
 & 7.1 & 6.9 & 6.9 \\

\specialrule{1.5pt}{\aboverulesep}{\belowrulesep}

\end{tabular}
}
\label{tab:ablation_lambda2}
\end{table}

\subsection{Training Efficiency vs. Robustness: A Bernoulli Mixing Analysis}
\label{appendix:BernoulliMixingReason}

We adopt Bernoulli mixing primarily for efficiency and effectiveness considerations. Using both perturbation types, $\delta_{\text{base}}$ and $\delta_{\text{hybrid}}$, at every iteration would require generating two adversarial examples per step, nearly doubling the training cost while yielding only marginal robustness gains.

To validate this design choice, we conduct an ablation study in which both perturbations are applied at every iteration (``Always Both'') and compare it against Bernoulli mixing. The results indicate that this setting leads to only slight improvements in robustness, while almost doubling the overall training time. This confirms that Bernoulli mixing provides a better trade-off between computational efficiency and adversarial robustness.

\paragraph{Experimental Setup.}
In Table \ref{tab:bernoulli-mixing-ablation}, we evaluate Whisper-Medium on Polish using $\approx$30 hours of audio. Models are trained with a batch size of 4 for 4 epochs, using the hyperparameters described in the main paper, and mixed-precision training on a single NVIDIA RTX-4090 (24GB).

\begin{table}[ht]
\centering
\caption{Ablation study comparing Bernoulli mixing and always applying both perturbations on Whisper‑Medium for Polish under clean and transferred adversarial settings ($\approx$30 hours of audio, batch size 4, 4 epochs).}
\resizebox{0.6\linewidth}{!}{%
\begin{tabular}{l c c c}
\specialrule{1.5pt}{\aboverulesep}{\belowrulesep}

\textbf{Setting} & \textbf{Clean WER/CER} & \textbf{Base-Adv WER/CER} & \textbf{Time (Hours)} \\

\specialrule{1.5pt}{\aboverulesep}{\belowrulesep}

Bernoulli Mixing & 12.0 / 3.8 & 24.2 / 9.7 & 20.8 \\
Always Both       & 12.2 / 3.8 & 23.8 / 9.5 & 40.5 \\

\specialrule{1.5pt}{\aboverulesep}{\belowrulesep}
\end{tabular}
}
\label{tab:bernoulli-mixing-ablation}
\end{table}

\subsection{Perceptual Analysis of Adversarial Audio}\label{appendix:perceptual_analysis}

We analyze the perceptual quality of $\approx$4 hours of adversarial samples to quantify how closely they preserve the characteristics of clean speech. All adversarial examples are generated under a strict 27 dB SNR constraint (perturbation energy $\approx 1/500$ of clean speech). Our generation procedure produces temporally uniform noise, avoiding narrow-band or impulsive distortions.
 
A perceptual analysis on $\approx$4 hours of adversarial examples from the Polish Common Voice dataset yields:

\begin{itemize}
    \item \textbf{STOI = 0.97}~\cite{taal2010stoi} (range [0, 1], higher is better): Indicates near-perfect speech intelligibility and preserved phonetic content.
    
    \item \textbf{Mean CV = 0.05}: The low coefficient of variation in frame energy confirms that the perturbation is evenly distributed over time, appearing as a steady background hiss rather than a jarring distortion.
    
    \item \textbf{PESQ = 3.0}~\cite{rix2001pesq} (range $\sim$[-0.5, 4.5], higher is better): Reflects fair-to-good perceptual quality. The slight degradation is mainly due to low-amplitude uniform background noise persisting during silent intervals between words, rather than degradation of the speech signal itself.
\end{itemize}

Overall, these results indicate that the perturbations may be faintly audible during pauses, but remain unobtrusive, highly intelligible, and free of severe artifacts.

\subsection{Effect of the HybridGrad Ratio $\gamma$}\label{subsec:gamma_ablation}

This subsection examines the effect of the gradient mixing ratio $\gamma$ in the HybridGrad-AFT algorithm. The parameter $\gamma$ controls the relative contribution of gradients from the base and target models, with smaller values emphasizing the base model and larger values increasing reliance on the target model.

To evaluate the sensitivity of HybridGrad-AFT to this parameter, we conduct an ablation study by varying $\gamma \in \{0.2, 0.5, 1.0\}$. Experiments are performed on Polish using Whisper-Medium and Wav2Vec2-Large-XLSR backbones. We report performance under three evaluation settings: clean inputs, Base-Adv, and Surrogate-Adv. The results are summarized in Table~\ref{tab:gamma_ablation}. Overall, $\gamma = 0.5$ yields the best trade-off across evaluation settings and is therefore used as the default in all other experiments.
\begin{table}[ht]
\centering
\caption{Effect of the HybridGrad ratio $\gamma$ on clean accuracy and adversarial transferability. Results are reported for Polish using Whisper-Medium and Wav2Vec2-Large-XLSR backbones. Lower is better.}
\resizebox{0.7\linewidth}{!}{%
\begin{tabular}{BBB c
  c c c
  c c c
}

\specialrule{1.5pt}{\aboverulesep}{\belowrulesep}

 &  &  &
 & \multicolumn{3}{c}{\textbf{Whisper Medium (Ours)}}
 & \multicolumn{3}{c}{\textbf{Wav2Vec2 Large XLSR (Ours)}} \\
\cmidrule(lr){5-7}\cmidrule(lr){8-10}

Language & Setup & Metric
 &
 & $\gamma = 0.2$ & $\gamma = 0.5$ & $\gamma = 1.0$
 & $\gamma = 0.2$ & $\gamma = 0.5$ & $\gamma = 1.0$ \\

\specialrule{1.5pt}{\aboverulesep}{\belowrulesep}

\multicolumn{1}{c}{\multirow{6}{*}{\textbf{Polish}}}
 & \multirow{2}{*}{Clean} & WER
 &  & 11.9 & 12.0 & 12.3
 & 16.2 & 16.8 & 16.7 \\

 &  & CER
 &  & 3.7 & 3.8 & 4.0
 & 4.0 & 4.2 & 4.2 \\

\cmidrule(lr){2-3}\cmidrule(lr){5-7}\cmidrule(lr){8-10}

 & \multirow{2}{*}{Base-Adv} & WER
 &  & 23.5 & 24.2 & 25.4
 & 24.3 & 24.8 & 25.0 \\

 &  & CER
 &  & 8.9 & 9.7 & 10.1
 & 6.8 & 6.9 & 7.0 \\

\cmidrule(lr){2-3}\cmidrule(lr){5-7}\cmidrule(lr){8-10}

 & \multirow{2}{*}{Surrogate-Adv} & WER
 &  & 48.5 & 44.1 & 45.2
 & 45.1 & 40.7 & 41.1 \\

 &  & CER
 &  & 22.7 & 20.5 & 21.2
 & 16.9 & 15.3 & 15.5 \\

\specialrule{1.5pt}{\aboverulesep}{\belowrulesep}

\end{tabular}
}
\label{tab:gamma_ablation}
\end{table}

\subsection{HybridGrad Perturbation Effectiveness}\label{appendix:hybridgrad_effectiveness}
We evaluate the effectiveness of the proposed HybridGrad attack, which is designed to be simultaneously effective on both the base and target models. Table~\ref{tab:mix_attack} reports the performance of the base model and Base-AFT target model under HybridGrad perturbations. These perturbations are generated by jointly using the base model and the Base-AFT target model, following the HybridGrad procedure. The attack is effective on both models, while the Base-AFT target model is more robust due to adversarial training.
\begin{table}[ht]
\centering
\caption{WER/CER of base and Base-AFT models under the HybridGrad adaptive attack on Polish and Portuguese, evaluated across ASR backbones. Higher = tronger attack.}
\resizebox{0.75\linewidth}{!}{%
\begin{tabular}{BBB c
  c c
  c c
  c c
}

\specialrule{1.5pt}{\aboverulesep}{\belowrulesep}

 &  &  &
 & \multicolumn{2}{c}{\textbf{Whisper Medium}}
 & \multicolumn{2}{c}{\textbf{Wav2Vec2 Large XLSR}}
 & \multicolumn{2}{c}{\textbf{MMS 1B}} \\
\cmidrule(lr){5-6}\cmidrule(lr){7-8}\cmidrule(lr){9-10}
Language & Setup & Metric
 &
 & \textbf{Base} & \textbf{Base-AFT}
 & \textbf{Base} & \textbf{Base-AFT}
 & \textbf{Base} & \textbf{Base-AFT} \\

\specialrule{1.5pt}{\aboverulesep}{\belowrulesep}

\multicolumn{1}{c}{\multirow{2}{*}{\textbf{Polish}}}
 & \multirow{2}{*}{Adv (HybridGrad)} & WER &  & 70.0 & 63.3 & 69.7 & 57.6 & 71.9 & 50.3 \\
 &  & CER &  & 40.4 & 28.9 & 32.1 & 26.5 & 34.2 & 20.8 \\

\specialrule{1.5pt}{\aboverulesep}{\belowrulesep}

\multicolumn{1}{c}{\multirow{2}{*}{\textbf{Portuguese}}}
 & \multirow{2}{*}{Adv (HybridGrad)} & WER &  & 75.6 & 62.5 & 85.2 & 64.4 & 84.5 & 62.7 \\
 &  & CER &  & 35.9 & 29.9 & 42.4 & 30.6 & 40.7 & 28.8 \\

\specialrule{1.5pt}{\aboverulesep}{\belowrulesep}

\end{tabular}
}
\label{tab:mix_attack}
\end{table}

\subsection{HybridGrad Scheduler: Design and Empirical Analysis}\label{appendix:hybridgrad_scheduler}

The proposed scheduler is designed to optimize gradient combination by explicitly accounting for the shift in the adversarial subspace of the surrogate model relative to the base and target models. Starting from the base model, the attacker fine-tunes it to construct a surrogate model, which progressively moves closer to the target model in representation space. As a result, the overlap between the surrogate and target adversarial subspaces increases during this process.

Motivated by this observation, the scheduler prioritizes the base-model gradient in the early PGD steps to maintain strong attack effectiveness on the base model. As optimization proceeds, it gradually shifts the weighting toward the target-model gradient to enhance transferability. This dynamic weighting strategy enables HybridGrad to generate perturbations that remain effective on both models.

\paragraph{Ablation Study.}
Table~\ref{tab:scheduler_ablation} presents a comparison between the proposed scheduler and two baselines: (i) Uniform weighting (0.5/0.5) and (ii) a Reverse scheduler (Target $\rightarrow$ Base). As shown, our scheduler generates stronger training-time perturbations, which in turn yield a more robustly trained model, achieving lower clean and adversarial WER/CER and thereby improving robustness under adaptive evaluation.












\begin{table}[ht]
\centering
\caption{Comparison of different HybridGrad scheduling strategies evaluated under clean settings, base-model adversarial attacks (Base-Adv), and surrogate-based adaptive adversarial attacks (Surrogate-Adv) on Whisper-Medium. Lower is better.}
\resizebox{0.52\linewidth}{!}{%
\begin{tabular}{l l c c c >{\columncolor[HTML]{e5e5ff}}c}

\specialrule{1.5pt}{\aboverulesep}{\belowrulesep}

& & & \multicolumn{3}{c}{\textbf{Whisper Medium}} \\ 
\cmidrule(lr){4-6}                                    

\textbf{Language} & \textbf{Setup} & \textbf{Metric} & \textbf{Uniform} & \textbf{Reverse} & \textbf{Ours} \\

\specialrule{1.5pt}{\aboverulesep}{\belowrulesep}

\multicolumn{1}{c}{\multirow{6}{*}{\textbf{Polish}}}
 & \multirow{2}{*}{\textcolor{gray}{\textbf{Clean}}} & \textcolor{gray}{\textbf{WER}} & \textcolor{gray}{12.7} & \textcolor{gray}{12.5} & \textcolor{gray}{12.0} \\
 &                        & \textcolor{gray}{\textbf{CER}} & \textcolor{gray}{4.0}  & \textcolor{gray}{4.0}  & \textcolor{gray}{3.8}  \\

\cmidrule(lr){2-6}

 & \multirow{2}{*}{\textbf{Base-Adv}}  & \textbf{WER} & 27.2 & 26.4 & \textbf{24.2} \\
 &                        & \textbf{CER} & 11.9 & 11.5 & \textbf{9.7}  \\

\cmidrule(lr){2-6}

 & \multirow{2}{*}{\textbf{Surrogate-Adv}} & \textbf{WER} & 49.0 & 44.8 & \textbf{44.1} \\
 &                                & \textbf{CER} & 23.7 & 20.9 & \textbf{20.5} \\

\specialrule{1.5pt}{\aboverulesep}{\belowrulesep}

\end{tabular}
}
\label{tab:scheduler_ablation}
\end{table}

\subsection{Robustness to Common Acoustic Corruptions}
\label{appendix:corruption_robustness}

We evaluate robustness to two common non-adversarial corruptions on the same Polish Common Voice test set used in the main experiments: MUSAN-derived babble noise~\cite{snyder2015musan} mixed at 15~dB SNR and AMR-NB compression at 6.7~kbps. We compare ordinarily fine-tuned (FT) and TransferBreaker models across three ASR backbones.

Under MUSAN-derived babble noise, TransferBreaker exhibits smaller degradation across all three backbones, with WER increases of 1.2--3.9 points compared with 3.3--10.4 points for FT. Whisper-Medium is the only case where TransferBreaker has slightly higher absolute corrupted WER (13.2 versus 12.9), reflecting its higher clean WER (12.0 versus 9.6), while its degradation remains smaller (1.2 versus 3.3 points). Under AMR-NB compression, TransferBreaker achieves both lower absolute WER and smaller degradation across all three backbones. These results suggest that its robustness partially extends to common acoustic corruptions, while remaining primarily targeted at the adversarial transfer threat model. The full results are reported in Table~\ref{tab:corruption_robustness}.

\begin{table}[ht]
\centering
\caption{WER (\%) under common non-adversarial acoustic corruptions on the Polish Common Voice test set. MUSAN-derived babble noise is mixed at 15~dB SNR, and AMR-NB compression uses the 6.7~kbps mode. Lower is better.}
\label{tab:corruption_robustness}
\resizebox{0.7\linewidth}{!}{%
\begin{tabular}{B c c c c c c}
\specialrule{1.5pt}{\aboverulesep}{\belowrulesep}
& \multicolumn{3}{c}{\textbf{FT}}
& \multicolumn{3}{c}{\textbf{Ours}} \\
\cmidrule(lr){2-4}\cmidrule(lr){5-7}

Model & Clean & MUSAN & AMR-NB & Clean & MUSAN & AMR-NB \\

\specialrule{1.5pt}{\aboverulesep}{\belowrulesep}

Whisper-Medium
& 9.6 & 12.9 & 17.8
& 12.0 & 13.2 & 15.6 \\

Wav2Vec2 Large XLSR
& 12.2 & 22.6 & 29.1
& 16.8 & 20.7 & 25.1 \\

MMS-1B
& 8.9 & 15.6 & 20.1
& 10.7 & 14.1 & 17.2 \\

\specialrule{1.5pt}{\aboverulesep}{\belowrulesep}
\end{tabular}
}
\end{table}

\section{Evaluation Metrics}\label{Appendix:Evaluation_Metrics}

We evaluate automatic speech recognition (ASR) performance using two standard metrics: Word Error Rate (WER) and Character Error Rate (CER). Both metrics are computed using the Levenshtein distance, measuring the normalized edit distance between a model’s predicted transcription and the corresponding ground-truth reference.

\subsection{Word Error Rate (WER)}

Word Error Rate (WER) assesses transcription accuracy at the word level. It is defined as the normalized Levenshtein distance between the predicted and reference word sequences, where the allowed edit operations include substitutions (\textit{S}), deletions (\textit{D}), and insertions (\textit{I}). Formally, WER is computed as
\[
\mathrm{WER} = \frac{S + D + I}{N},
\]
where $N$ denotes the total number of words in the reference transcription.

\subsection{Character Error Rate (CER)}

Character Error Rate (CER) evaluates transcription accuracy at the character level. Analogous to WER, CER is computed as the normalized Levenshtein distance between the predicted and reference character sequences, with character-level substitutions ($S_c$), deletions ($D_c$), and insertions ($I_c$) as the permitted edit operations. CER is defined as
\[
\mathrm{CER} = \frac{S_c + D_c + I_c}{N_c},
\]
where $N_c$ represents the total number of characters in the reference transcription.



\section{Supplementary Implementation Details}\label{Appendix:Extra_Implemetation_Detail}
During training, adversarial examples are generated with $n=10$ steps, a step size $\alpha_x=0.25\epsilon$, and momentum coefficient $\mu=0.9$, where the perturbation budget $\epsilon$ is specified by a signal-to-noise ratio (SNR) constraint of $22~\mathrm{dB}$. 
We use a stronger (lower-SNR) budget during training than at evaluation because strict imperceptibility is not required for training-time perturbations as long as clean performance is preserved, and the larger budget compensates for the comparatively efficient, 10-step PGD attack used to generate them.
For evaluation, we use our \textit{Advance-PGD} (see Appendix~\ref{appendix:AdvancePGD}) with $n=50$ steps and an SNR constraint of $27~\mathrm{dB}$, matching the perceptual constraint used in our threat model; our perceptual metrics (Appendix~\ref{appendix:perceptual_analysis}) indicate that adversarial examples generated at 27~dB remain highly intelligible and free of severe artifacts. Notably, TransferBreaker is not tied to either specific budget: Figure~\ref{fig:snr_figure} shows it remains robust across a wide range of SNR constraints (17--37~dB). All baseline defenses in our comparisons share the same training and evaluation configurations and hyperparameters (e.g., Std-AFT is also trained and evaluated under the same hyperparameters). Attack comparisons use the same evaluation perturbation budget, with method-specific optimization settings described in Appendix~\ref{subsec:other_attacks}. The surrogate model for adaptive attacks in our framework is obtained by applying the same training procedure to the MLS dataset, a large-scale multilingual corpus for automatic speech recognition. For LJ-Regularization, we apply the regularization term to the encoder outputs across all model architectures.




\subsection{Advance-PGD}\label{appendix:AdvancePGD}
For robustness evaluation, we employ a customized PGD-based adversarial attack, referred to as \textit{Advance-PGD}. The core optimization procedure follows the same principle as Algorithm~\ref{alg:delta_base}, performing iterative gradient-based updates with momentum. In contrast to Algorithm~\ref{alg:delta_base}, Advance-PGD incorporates an adaptive step size $\alpha$, which is dynamically adjusted based on the observed loss trajectory during optimization. Specifically, $\alpha$ is adapted based on loss stagnation: if the loss does not improve for 10 consecutive steps, $\alpha$ is multiplied by 0.5. The stagnation counter is reset whenever an improvement in the loss is observed. This adaptive strategy improves the stability of adversarial example generation and enables finer-grained perturbation refinement, particularly in later iterations.

\section{Training Computational Cost Analysis}\label{appendix:training_cost_comparison}

Table~\ref{tab:training_cost_comparison} reports training time and memory usage for normal fine-tuning (FT), standard adversarial fine-tuning (Std-AFT), and TransferBreaker, using Std-AFT as the primary baseline.

\textbf{Setup.} All experiments are conducted on a single RTX-4090 GPU (24 GB) using mixed precision, over approximately 30 hours of Polish utterances. Batch sizes are 4 for Whisper-Medium and 8 for Wav2Vec2-Large-XLSR.

\begin{table}[H]
\centering
\caption{Training cost comparison between normal fine-tuning (FT), standard adversarial fine-tuning (Std-AFT), and TransferBreaker using Whisper-Medium and Wav2Vec2-Large-XLSR. Experiments are conducted on a single RTX-4090 (24 GB) using $\approx$30 hours of Polish audio.}
\resizebox{0.6\linewidth}{!}{%
\begin{tabular}{l
  c c c
  c c c
}

\specialrule{1.5pt}{\aboverulesep}{\belowrulesep}

& \multicolumn{3}{c}{\textbf{Whisper Medium}} 
& \multicolumn{3}{c}{\textbf{Wav2Vec2 Large XLSR}} \\

\cmidrule(lr){2-4}\cmidrule(lr){5-7}

\textbf{Metric}
& \textbf{FT} & \textbf{Std-AFT} & \textbf{Ours}
& \textbf{FT} & \textbf{Std-AFT} & \textbf{Ours} \\

\specialrule{1.5pt}{\aboverulesep}{\belowrulesep}

\textbf{Peak VRAM (GB)}
& 17.2 & 18.9 & 21.7
& 18.1 & 18.4 & 19.3 \\

\textbf{Time / Epoch (Hours)}
& 0.42 & 2.91 & 5.20
& 0.14 & 1.00 & 1.80 \\

\specialrule{1.5pt}{\aboverulesep}{\belowrulesep}

\end{tabular}
}
\label{tab:training_cost_comparison}
\end{table}

\textbf{Overhead relative to Std-AFT.} Since TransferBreaker builds on adversarial fine-tuning, we report its overhead relative to Std-AFT. It introduces approximately a 1.8$\times$ increase in training time per epoch for both models, along with a modest VRAM increase of +2.8 GB (Whisper) and +0.9 GB (Wav2Vec2). Despite maintaining a frozen base model in memory, peak usage remains within a single consumer GPU.

\textbf{Practicality.} In a typical fine-tuning setting with moderately resourced languages, this overhead remains manageable in terms of wall-clock time. The frozen base model introduces no optimizer state, and the additional cost occurs only during training, with no impact on inference.

Overall, TransferBreaker introduces a moderate and predictable overhead over Std-AFT, making it practical for adaptation to moderately resourced languages.

\section{Interpreting Adaptive-Attack Results}
\label{appendix:TradeOffAccRob}

As shown in Table~\ref{tab:aft_variants}, TransferBreaker improves robustness over Base-AFT and Base+Std-AFT under adaptive attacks. We additionally evaluate Standard-AFT under the same Surrogate-Adv setting; averaged across the six Polish and Portuguese model--language settings, its WER is 45.0.

Adversarial robustness remains an open challenge, even in extensively studied domains such as ImageNet. For example, while the best standard models achieve around 87.76\% clean accuracy \cite{Liu_2022_CVPR}, the best robust models achieve approximately 77.00\% clean accuracy and 57.70\% robust accuracy~\cite{singh2023revisiting}, leaving a substantial robustness gap. This highlights the well-known clean-accuracy/robustness trade-off in adversarial robustness. Adaptive attacks are particularly challenging because they are close to white-box attacks, which remain difficult even in mature classification benchmarks. Therefore, the relatively high adaptive-attack success rate should not be interpreted as a limitation specific to TransferBreaker, but rather as a reflection of the broader open challenge of defending against near-white-box adversaries.

Importantly, TransferBreaker preserves strong clean performance while reducing transferability. Specifically, it achieves a clean WER of 18.9, compared to 16.4 for clean fine-tuning, while substantially improving robustness relative to existing defenses. This behavior is consistent with the well-known clean-performance/robustness trade-off in adversarial robustness, as discussed in the main paper: TransferBreaker improves adversarial robustness without excessive degradation in clean performance~\cite{pmlr-v119-raghunathan20a,zhang2019theoretically,tsipras2019robustnessoddsaccuracy,nafez2025frameshield,mirzaei2025adversarially}.

Moreover, the adaptive-attack setting is nearly white-box, where Standard-AFT is expected to perform best. Nevertheless, TransferBreaker still improves average WER under the adaptive attack from 45.0 for Standard-AFT to 43.1. The gains are even larger under the base attack, where WER improves from 42.8 to 27.8. These results show that TransferBreaker substantially reduces transferability in the base-attack setting and still provides improvements in the near-white-box adaptive setting, where Standard-AFT would otherwise be expected to have an advantage.

\section{White-Box Robustness on Fine-Tuned Targets}\label{appendix:whitebox_on_target}

Base-AFT is trained using perturbations generated on the base model, matching the base-to-target transfer threat model of Section~\ref{sec:threat_model}. We separately evaluate the adaptive setting, in which the attacker crafts perturbations using a surrogate model, in Appendix~\ref{appendix:surrogate_fidelity}. This section instead evaluates a stronger setting in which perturbations are generated directly on each fine-tuned target model, using Advance-PGD (Appendix~\ref{appendix:AdvancePGD}) under the same 27~dB SNR constraint, with gradients computed on the target itself. We compare Base-AFT, Standard-AFT, and TransferBreaker under this evaluation, reported in Table~\ref{tab:whitebox_on_target}, which clarifies whether TransferBreaker's robustness is specific to base-driven directions or reflects broader adversarial robustness.

\begin{table}[H]
\centering
\caption{WER (\%) under a white-box attack generated directly on each fine-tuned target model trained using Base-AFT, Std-AFT, and TransferBreaker. Lower indicates stronger robustness.}
\resizebox{0.6\linewidth}{!}{%
\begin{tabular}{BB cc c}
\specialrule{1.5pt}{\aboverulesep}{\belowrulesep}
 &  & \multicolumn{3}{c}{\textbf{Target Defended By}} \\
\cmidrule(lr){3-5}
Language & Model & \textbf{Base-AFT} & \textbf{Std-AFT} & \textbf{TransferBreaker} \\
\specialrule{1.5pt}{\aboverulesep}{\belowrulesep}
\multirow{2}{*}{Polish} & Whisper-Medium & 65.8 & 57.0 & 57.5 \\
 & Wav2Vec2 & 61.9 & 54.1 & 54.5 \\
\cmidrule(lr){1-5}
\multirow{2}{*}{Portuguese} & Whisper-Medium & 64.7 & 57.8 & 58.3 \\
 & Wav2Vec2 & 69.8 & 61.6 & 61.0 \\
\specialrule{1.5pt}{\aboverulesep}{\belowrulesep}
\end{tabular}
}
\label{tab:whitebox_on_target}
\end{table}

\textbf{TransferBreaker generalizes to broad white-box robustness.} According to Table~\ref{tab:whitebox_on_target}, TransferBreaker matches Standard-AFT's white-box performance within $\pm0.6$ WER across all four settings, the regime for which Standard-AFT's training objective is explicitly designed (Section~\ref{Base_Theory_SubSubSection}), and both substantially outperform the undefended fine-tuned lower bound of 84 to 110 WER (Table~\ref{tab:main_results}). Moreover, TransferBreaker decisively outperforms Standard-AFT on base-model transfer attacks ($27.8$ versus $42.8$ average WER, Table~\ref{tab:main_results}) and on adaptive attacks ($43.1$ versus $45.0$ WER for Standard-AFT, Section~\ref{appendix:TradeOffAccRob}). We attribute this to HybridGrad-AFT focusing training on the fundamental perturbation subspace $V_F$ (Appendix~\ref{HybridGrad-AFT}), which is shared across models and therefore covers many of the directions a white-box attacker would otherwise exploit.

\textbf{Base-AFT alone is narrower.} Its weaker white-box robustness mirrors the vulnerability to adaptive attacks reported in Table~\ref{tab:aft_variants}, motivating HybridGrad-AFT's design (Section~\ref{sec:adaptive_thread_model}).

\textbf{TransferBreaker's gains are not an artifact of matched base-derived perturbations in training and evaluation.} Robustness persists under perturbations generated directly on the target, where TransferBreaker effectively matches Standard-AFT, remaining within 0.6 WER across all four settings. The high absolute WER at this near-white-box extreme reflects the accuracy-robustness trade-off and the open challenge of white-box adversarial robustness discussed in Section~\ref{appendix:TradeOffAccRob}, amplified in ASR because WER aggregates errors across full output sequences.

\section{Surrogate Fidelity in the Adaptive Threat Model}\label{appendix:surrogate_fidelity}

The adaptive attacker of Section~\ref{sec:adaptive_thread_model} trains its surrogate model on the Multilingual LibriSpeech (MLS) dataset, which differs from the target's private training distribution. Common Voice, used to fine-tune the target, consists of crowdsourced, prompted speech from many speakers under heterogeneous recording conditions, whereas MLS consists of read audiobook speech from fewer speakers with cleaner acoustics. This setup reflects a realistic strong adversary who holds the defense procedure and access to abundant public data, but not the private training set.

Rather than characterizing the degree of similarity between these two corpora, we evaluate the limiting case of zero distributional mismatch, which subsumes any weaker degree of similarity: an additional surrogate trained on the same Common Voice data used for the target, differing only in training randomness (random seed). We compare this same-data surrogate against the base-model attack of
Table~\ref{tab:main_results}, the MLS-based adaptive surrogate of
Table~\ref{tab:aft_variants}, and a full white-box attack on the private
target, trained using TransferBreaker and Std-AFT (which is optimized for
the white-box setting), from Appendix~\ref{appendix:whitebox_on_target}.
Table~\ref{tab:surrogate_fidelity_spectrum} reports the resulting WER
across this spectrum of attacker knowledge for Polish and Portuguese, using
Whisper-Medium and Wav2Vec2.

Even the same-data surrogate, the closest possible approximation of the target short of direct parameter access, remains approximately $8$ to $9$ WER points below the white-box ceiling (TransferBreaker target) across all four settings. This indicates that surrogate-based attacks are bounded away from white-box strength by divergence in the optimization trajectory itself, rather than by dataset mismatch alone. As discussed in Appendix~\ref{HybridGrad-AFT}, successful surrogate transfer requires perturbations to remain effective across distinct models. Divergence in their optimization trajectories can therefore reduce the overlap between their adversarial perturbation sets even when the training data are matched, while fundamental transfer directions remain shared across models.

Consequently, the adaptive threat model of Section~\ref{sec:adaptive_thread_model} approaches, but does not collapse into, a standard white-box scenario, even when the attacker holds the exact defense procedure, training data, architecture, initialization, and hyperparameters used by the target. Both endpoints of this spectrum, an attacker with no target-specific knowledge (no access to the target's parameters or training data) and one with full parameter access, represent idealized extremes that bracket practical private-deployment threat models; TransferBreaker remains the strongest defense across the more realistic settings between them, improving average WER over Standard-AFT under both base-model attacks ($27.8$ versus $42.8$, Table~\ref{tab:main_results}) and the evaluated adaptive attacks ($43.1$ versus $45.0$, Section~\ref{appendix:TradeOffAccRob}).

\begin{table}[H]
\centering
\caption{Attack strength against TransferBreaker across the surrogate-fidelity spectrum, from a base-model attack to a full white-box attack on the target. Values are WER (\%); higher indicates a stronger attack. Same-data surrogate (seed only) is shown for reference.}
\resizebox{\linewidth}{!}{%
\begin{tabular}{BB cccc}
\specialrule{1.5pt}{\aboverulesep}{\belowrulesep}
 &  & \multicolumn{2}{c}{\textbf{Polish}} & \multicolumn{2}{c}{\textbf{Portuguese}} \\
\cmidrule(lr){3-4}\cmidrule(lr){5-6}
Knowledge & Surrogate Model & Whisper-Medium & Wav2Vec2 & Whisper-Medium & Wav2Vec2 \\
\specialrule{1.5pt}{\aboverulesep}{\belowrulesep}
Base model only & None (base model) & 24.2 & 24.8 & 21.7 & 32.7 \\
\cmidrule(lr){1-6}
Full white-box & TransferBreaker target (ceiling) & 57.5 & 54.5 & 58.3 & 61.0 \\
\cmidrule(lr){1-6}
Full white-box & Std-AFT target (reference) & 57.0 & 54.1 & 57.8 & 61.6 \\
\cmidrule(lr){1-6}
Defense procedure & MLS surrogate & 44.1 & 40.7 & 44.7 & 48.9 \\
\cmidrule(lr){1-6}
\rowcolor[HTML]{e5e5ff} Defense procedure + target data & Same-data surrogate (seed only) & 48.6 & 45.2 & 49.1 & 53.4 \\
\specialrule{1.5pt}{\aboverulesep}{\belowrulesep}
\end{tabular}
}
\label{tab:surrogate_fidelity_spectrum}
\end{table}

\section{Persistent Suboptimality of Standard-AFT Across Epochs}\label{appendix:std-aft-across-epochs}

While Standard-AFT seems logical in a full white-box setting, Table \ref{tab:main_results} shows that it remains suboptimal under the base-to-descendant transfer threat model and does not match the performance of Base-AFT. As argued in Section \ref{section:Base-Adversarial-Fine-Tuning}, the optimal defense should specifically neutralize base-derived attacks. Section \ref{Base_Theory_SubSubSection} shows that Base-AFT concentrates robustness on the structured subspace where the base model is most vulnerable, rather than across the entire perturbation space. In contrast, Standard-AFT’s objective is overly broad for complex ASR tasks and may over-constrain the model. Base-AFT provides a more efficient defense by aligning directly with the attacker’s actual capabilities in this setting.

\textbf{Ablation Study.} To examine whether the performance limitation of Standard-AFT can be mitigated with additional fine-tuning, we double the number of epochs from 4 to 8 for both Whisper-Medium and Wav2Vec2-Large-XLSR on the Polish Common Voice dataset (Table \ref{tab:epoch_ablation_std_aft}). The results show only marginal improvements, with performance still lagging behind Base-AFT fine-tuned for 4 epochs (e.g., Base-Adv WER for Whisper: 41.5 vs. 22.4; for Wav2Vec2: 31.4 vs. 24.0). This indicates that Standard-AFT does not subsume Base-AFT, even with extended optimization.















\begin{table}[ht]
\centering
\caption{Effect of increasing training epochs for Standard‑AFT on Whisper‑Medium and Wav2Vec2‑Large‑XLSR for Polish under clean and transferred adversarial settings. Base-AFT (4 epochs) is shown for comparison. Lower is better.}
\resizebox{0.85\linewidth}{!}{%
\begin{tabular}{c c c c c c >{\columncolor[HTML]{e5e5ff}}c c c >{\columncolor[HTML]{e5e5ff}}c}
\specialrule{1.5pt}{\aboverulesep}{\belowrulesep}
 &  &  & & \multicolumn{3}{c}{\textbf{Whisper Medium}} & \multicolumn{3}{c}{\textbf{Wav2Vec2 Large XLSR}} \\
\cmidrule(lr){5-7}\cmidrule(lr){8-10}
\textbf{Language} & \textbf{Setup} & \textbf{Metric} & & \textbf{Std-AFT(4e)} & \textbf{Std-AFT(8e)} & \textbf{Base-AFT(4e)} & \textbf{Std-AFT(4e)} & \textbf{Std-AFT(8e)} & \textbf{Base-AFT(4e)} \\
\specialrule{1.5pt}{\aboverulesep}{\belowrulesep}
\multirow{4}{*}{\textbf{Polish}}
 & \multirow{2}{*}{\textbf{\color{CG}Clean}} & \textbf{\color{CG}WER} & & \color{CG}12.9 & \color{CG}12.5 & \color{CG}11.0 & \color{CG}17.3 & \color{CG}17.0 & \color{CG}15.9 \\
 &                        & \textbf{\color{CG}CER} & & \color{CG}4.2  & \color{CG}4.1  & \color{CG}3.4  & \color{CG}4.5  & \color{CG}4.4  & \color{CG}3.9  \\
\cmidrule(lr){2-3}\cmidrule(lr){5-7}\cmidrule(lr){8-10}
 & \multirow{2}{*}{\textbf{Base-Adv}}   & \textbf{WER} & & 44.0 & 41.5 & \textbf{22.4} & 33.7 & 31.4 & \textbf{24.0} \\
 &                        & \textbf{CER} & & 22.5 & 21.3 & \textbf{8.3}  & 10.3 & 9.8  & \textbf{6.9}  \\
\specialrule{1.5pt}{\aboverulesep}{\belowrulesep}
\end{tabular}
}
\label{tab:epoch_ablation_std_aft}
\end{table}

\section{Comparison with Preprocessing-Based Defense}
\label{appendix:PreprocessingBaseDefence}



For additional comparison, we evaluate two preprocessing-based approaches: (i) a diffusion-based purification method proposed by Kühne et al. \cite{DiffusionDefence}, implemented with their recommended settings on a fine-tuned model (Purifier+FT), and (ii) a fine-tuned model combined with a CleanUNet denoiser (CleanUNet+FT) \cite{kong2022cleanunet}. CleanUNet is a standard speech denoising model included to assess whether generic denoising can reduce adversarial perturbations. Experiments are conducted using Whisper-Medium on Polish and Portuguese, comparing Standard-AFT, CleanUNet+FT, Purifier+FT, and our method under both clean and transferred-adversarial conditions.

Table \ref{tab:preprocessing_comparison} shows that diffusion purification substantially degrades clean accuracy; for example, on Portuguese, Purifier+FT increases clean WER to 29.4, compared to 15.1 for our method, making it impractical when most inputs are benign. It also offers limited robustness, with adversarial WER reaching 54.4 on Portuguese. CleanUNet+FT maintains clean accuracy comparable to fine-tuning alone, yet fails to mitigate transferred adversarial attacks, reaching an adversarial WER of 62.5 on Portuguese. In contrast, our method achieves markedly stronger robustness (21.7 adversarial WER on Portuguese) while maintaining competitive clean accuracy.










\begin{table}[H]
\centering
\caption{Comparison of diffusion purification with fine-tuning (Purifier+FT) and CleanUNet denoising with fine-tuning (CleanUNet+FT) against standard-AFT (Std-AFT) and our method on Whisper-Medium for Polish and Portuguese under clean and transferred adversarial settings. Lower is better.}
\resizebox{0.6\linewidth}{!}{%
\begin{tabular}{lll c c c c >{\columncolor[HTML]{e5e5ff}}c}  

\specialrule{1.5pt}{\aboverulesep}{\belowrulesep}

& & & & \multicolumn{4}{c}{\textbf{Whisper Medium}} \\ 
\cmidrule(lr){5-8}

\textbf{Language} & \textbf{Setup} & \textbf{Metric} & & \textbf{Std-AFT} & \textbf{CleanUNet+FT} & \textbf{Purifier+FT} & \textbf{Ours} \\

\specialrule{1.5pt}{\aboverulesep}{\belowrulesep}

\multicolumn{1}{c}{\multirow{4}{*}{\textbf{Polish}}} 
 & \multirow{2}{*}{\textbf{\color{CG}Clean}} & \textbf{\color{CG}WER} & & \color{CG}12.9 & \color{CG}10.9 & \color{CG}23.1 & \color{CG}12.0 \\
 &                        & \textbf{\color{CG}CER} & & \color{CG}4.2  & \color{CG}3.5 & \color{CG}11.2 & \color{CG}3.8  \\
\cmidrule(lr){2-3}\cmidrule(lr){5-8}                         
 & \multirow{2}{*}{\textbf{Base-Adv}}   & \textbf{WER} & & 44.0 & 64.2 & 51.8 & \textbf{24.2} \\
 &                        & \textbf{CER} & & 22.5 & 33.8 & 28.3 & \textbf{9.7}  \\

\specialrule{1pt}{\aboverulesep}{\belowrulesep}

\multicolumn{1}{c}{\multirow{4}{*}{\textbf{Portuguese}}} 
 & \multirow{2}{*}{\textbf{\color{CG}Clean}} & \textbf{\color{CG}WER} & & \color{CG}16.1 & \color{CG}14.2 & \color{CG}29.4 & \color{CG}15.1 \\
 &                        & \textbf{\color{CG}CER} & & \color{CG}5.5  & \color{CG}4.0 & \color{CG}13.9 & \color{CG}4.8  \\
\cmidrule(lr){2-3}\cmidrule(lr){5-8}                         
 & \multirow{2}{*}{\textbf{Base-Adv}}   & \textbf{WER} & & 42.9 & 62.5 & 54.4 & \textbf{21.7} \\
 &                        & \textbf{CER} & & 21.1 & 33.1 & 30.0 & \textbf{8.2}  \\

\specialrule{1.5pt}{\aboverulesep}{\belowrulesep}

\end{tabular}
}
\label{tab:preprocessing_comparison}
\end{table}

\section{Evaluation of TransferBreaker on extremely Low-resource Languages}\label{appendix:eval_extremly_low_resource}

We conduct additional experiments on Finnish, a Uralic language with highly agglutinative morphology, and Vietnamese, a tonal language in the Austroasiatic family, using the Common Voice 24 dataset. These settings are substantially more resource‑constrained than our primary experiments, with only approximately 2.5 hours and 2.3 hours of data in the predefined Common Voice training split for Finnish and Vietnamese, respectively. We further apply standard data augmentation, yielding an effective 4× expansion of the training set.

Despite the extremely limited training data, Table~\ref{tab:low_resource} shows that TransferBreaker remains highly effective in these low-resource settings, achieving substantial reductions in adversarial WER/CER while preserving competitive clean performance.













\begin{table}[H]
\centering
\caption{Evaluation of TransferBreaker on extremely low-resource languages (Finnish and Vietnamese) using Whisper-Medium on Common Voice 24 under clean and transferred adversarial settings. Lower is better.}
\resizebox{0.43\linewidth}{!}{%
\begin{tabular}{lll c c c >{\columncolor[HTML]{e5e5ff}}c}

\specialrule{1.5pt}{\aboverulesep}{\belowrulesep}

& & & & \multicolumn{3}{c}{\textbf{Whisper Medium}} \\
\cmidrule(lr){5-7}

\textbf{Language} & \textbf{Setup} & \textbf{Metric} & & \textbf{Base} & \textbf{FT} & \textbf{Ours} \\

\specialrule{1.5pt}{\aboverulesep}{\belowrulesep}

\multicolumn{1}{c}{\multirow{4}{*}{\textbf{Finnish}}}
 & \multirow{2}{*}{\textbf{\color{CG}Clean}} & \textbf{\color{CG}WER} & & \color{CG}19.8 & \color{CG}15.8 & \color{CG}17.6 \\
 &                        & \textbf{\color{CG}CER} & & \color{CG}6.8  & \color{CG}5.2  & \color{CG}5.6  \\
\cmidrule(lr){2-3}\cmidrule(lr){5-7}
 & \multirow{2}{*}{\textbf{Base-Adv}}   & \textbf{WER} & & 132.9 & 96.3 & \textbf{26.0} \\
 &                        & \textbf{CER} & & 82.8  & 51.4 & \textbf{10.1} \\

\specialrule{1pt}{\aboverulesep}{\belowrulesep}

\multicolumn{1}{c}{\multirow{4}{*}{\textbf{Vietnamese}}}
 & \multirow{2}{*}{\textbf{\color{CG}Clean}} & \textbf{\color{CG}WER} & & \color{CG}27.4 & \color{CG}20.7 & \color{CG}22.1 \\
 &                        & \textbf{\color{CG}CER} & & \color{CG}8.7  & \color{CG}6.2  & \color{CG}7.3  \\
\cmidrule(lr){2-3}\cmidrule(lr){5-7}
 & \multirow{2}{*}{\textbf{Base-Adv}}   & \textbf{WER} & & 157.7 & 94.3 & \textbf{34.6} \\
 &                        & \textbf{CER} & & 89.6  & 52.1 & \textbf{13.0} \\

\specialrule{1.5pt}{\aboverulesep}{\belowrulesep}

\end{tabular}
}
\label{tab:low_resource}
\end{table}

\section{Justification for the First‑Order Taylor Analysis}
\label{appendix:why_taylor_expansion}

Our theoretical analysis adopts a first-order perspective based on a Taylor expansion ``for small $\epsilon$'' which is a standard and widely used approach for explaining adversarial sensitivity and transferability, as it captures the dominant gradient/Jacobian directions governing local perturbation effects. While this linearization does not fully characterize arbitrarily large-$\epsilon$ behavior, prior work shows its practical relevance beyond the infinitesimal regime. In particular, \cite{simon2019firstorder} shows that although higher-order terms may dominate in principle for finite $\epsilon$, empirical evidence strongly suggests that first-order terms suffice to explain much of the observed adversarial vulnerability.

In our experiments, adversarial perturbations are constrained to a fixed 27\,dB SNR with uniformly distributed noise, which bounds perturbation energy and avoids extreme local distortions. Under this controlled, norm-bounded setup, perturbations remain within a regime where first-order effects dominate, making the Taylor approximation sufficient to explain the observed transferability trends.

\section{Motivation and Scope of Untargeted Attacks}\label{appendix:untargeted_scope}

Untargeted ASR degradation is practically relevant whenever an attacker seeks to make speech unreliable rather than force a specific transcript, enabling evasion of transcription, moderation, keyword search, audit logging, and compliance monitoring. We study digital, sample-specific, untargeted attacks as a strong, minimally constrained setting: digital attacks are applied directly to the input waveform without over-the-air transformations that can attenuate or distort the perturbation; sample-specific attacks optimize a separate perturbation for each utterance, whereas universal attacks reuse a single perturbation across inputs and therefore trade per-utterance effectiveness for scalability; and targeted attacks impose an additional constraint, as they must induce a specific, often lengthy output sequence. In principle, a sufficiently strong untargeted robustness guarantee would also preclude targeted attacks, since preserving the correct output against every alternative prevents any specific targeted output \cite{pmlr-v80-wong18a}. However, we do not provide such a guarantee or directly evaluate targeted attacks in this work.

\section{Characterizing Adversarial Error Modes Across ASR Architectures}\label{appendix:error_modes}

We examine whether adversarial WER inflation stems primarily from insertions
(hallucinated content) or from substitutions/deletions (lost content), and
find this is architecture-dependent.

\textbf{Autoregressive encoder-decoder (Whisper): insertion-dominated.}
Because the decoder conditions on its own previous tokens, a corrupted encoder representation drives it into self-reinforcing repetition loops or, when acoustic evidence weakens, into fluent but unrelated text sampled from
its language prior. Both failure modes produce unbounded insertions, explaining Whisper's WER exceeding 100\% in Table~\ref{tab:main_results} (e.g., 164.2\% on Arabic). For example:
\begin{quote}
\textit{Reference:} O comboio chegou \`a esta\c{c}\~ao com dez minutos de atraso. \\
\textit{Adversarial:} Legendas pela comunidade Amara.org. Obrigado por assistir.
\end{quote}
This prior-takeover mode is particularly safety-critical, as the output
transcript is fluent and carries no surface signal that an error occurred.

\textbf{CTC (Wav2Vec2-XLSR, MMS-1B): substitution/deletion-dominated.}
With one output distribution per frame, monotonic alignment, and no
autoregressive feedback, output length is frame-bounded and errors remain
local rather than open-ended. CTC models can still exceed 100\% WER (e.g.,
MMS-1B at 171.0\% on Portuguese), but through token fragmentation: a single
reference word is emitted as several malformed fragments, each scored as an
insertion, e.g.\ \textit{``zbiory by\l y obfite''} $\rightarrow$
\textit{``z biory b\l e o bfte''}. Unlike the decoder's repetition loops,
these errors stay bounded and localized to the corrupted region.


\newpage
\section*{NeurIPS Paper Checklist}

\begin{enumerate}

\item {\bf Claims}
    \item[] Question: Do the main claims made in the abstract and introduction accurately reflect the paper's contributions and scope?
    \item[] Answer: \answerYes{} 
    \item[] Justification: The abstract and introduction clearly state our specific claims and contributions, which are completely addressed, theoretically analyzed, and empirically validated in the subsequent sections of the paper.
    \item[] Guidelines:
    \begin{itemize}
        \item The answer \answerNA{} means that the abstract and introduction do not include the claims made in the paper.
        \item The abstract and/or introduction should clearly state the claims made, including the contributions made in the paper and important assumptions and limitations. A \answerNo{} or \answerNA{} answer to this question will not be perceived well by the reviewers. 
        \item The claims made should match theoretical and experimental results, and reflect how much the results can be expected to generalize to other settings. 
        \item It is fine to include aspirational goals as motivation as long as it is clear that these goals are not attained by the paper. 
    \end{itemize}

\item {\bf Limitations}
    \item[] Question: Does the paper discuss the limitations of the work performed by the authors?
    \item[] Answer: \answerYes{} 
    \item[] Justification: We explicitly discuss the clean-performance/robustness trade-off and the challenges of near-white-box adaptive attacks in Appendix~\ref{appendix:TradeOffAccRob}, as well as the scope limitation regarding targeted attacks in Appendix~\ref{appendix:untargeted_scope}.
    \item[] Guidelines:
    \begin{itemize}
        \item The answer \answerNA{} means that the paper has no limitation while the answer \answerNo{} means that the paper has limitations, but those are not discussed in the paper. 
        \item The authors are encouraged to create a separate ``Limitations'' section in their paper.
        \item The paper should point out any strong assumptions and how robust the results are to violations of these assumptions (e.g., independence assumptions, noiseless settings, model well-specification, asymptotic approximations only holding locally). The authors should reflect on how these assumptions might be violated in practice and what the implications would be.
        \item The authors should reflect on the scope of the claims made, e.g., if the approach was only tested on a few datasets or with a few runs. In general, empirical results often depend on implicit assumptions, which should be articulated.
        \item The authors should reflect on the factors that influence the performance of the approach. For example, a facial recognition algorithm may perform poorly when image resolution is low or images are taken in low lighting. Or a speech-to-text system might not be used reliably to provide closed captions for online lectures because it fails to handle technical jargon.
        \item The authors should discuss the computational efficiency of the proposed algorithms and how they scale with dataset size.
        \item If applicable, the authors should discuss possible limitations of their approach to address problems of privacy and fairness.
        \item While the authors might fear that complete honesty about limitations might be used by reviewers as grounds for rejection, a worse outcome might be that reviewers discover limitations that aren't acknowledged in the paper. The authors should use their best judgment and recognize that individual actions in favor of transparency play an important role in developing norms that preserve the integrity of the community. Reviewers will be specifically instructed to not penalize honesty concerning limitations.
    \end{itemize}

\item {\bf Theory assumptions and proofs}
    \item[] Question: For each theoretical result, does the paper provide the full set of assumptions and a complete (and correct) proof?
    \item[] Answer: \answerYes{} 
    \item[] Justification: We provide theoretical analyses and justifications for Base-AFT, Latent Jacobian Regularization, and HybridGrad-AFT in Sections~\ref{Base_Theory_SubSubSection}, \ref{LJR_theory}, and Appendix~\ref{HybridGrad-AFT}, with the scope of the first-order approximation discussed in Appendix~\ref{appendix:why_taylor_expansion}.
    \item[] Guidelines:
    \begin{itemize}
        \item The answer \answerNA{} means that the paper does not include theoretical results. 
        \item All the theorems, formulas, and proofs in the paper should be numbered and cross-referenced.
        \item All assumptions should be clearly stated or referenced in the statement of any theorems.
        \item The proofs can either appear in the main paper or the supplemental material, but if they appear in the supplemental material, the authors are encouraged to provide a short proof sketch to provide intuition. 
        \item Inversely, any informal proof provided in the core of the paper should be complemented by formal proofs provided in appendix or supplemental material.
        \item Theorems and Lemmas that the proof relies upon should be properly referenced. 
    \end{itemize}

    \item {\bf Experimental result reproducibility}
    \item[] Question: Does the paper fully disclose all the information needed to reproduce the main experimental results of the paper to the extent that it affects the main claims and/or conclusions of the paper (regardless of whether the code and data are provided or not)?
    \item[] Answer: \answerYes{} 
    \item[] Justification: We provide the code necessary to reproduce our results, alongside comprehensive implementation details and hyperparameters, in Section \ref{sec:experiments} and Appendices \ref{Appendix:Extra_Implemetation_Detail} and \ref{appendix_dataset_detail}.
    \item[] Guidelines:
    \begin{itemize}
        \item The answer \answerNA{} means that the paper does not include experiments.
        \item If the paper includes experiments, a \answerNo{} answer to this question will not be perceived well by the reviewers: Making the paper reproducible is important, regardless of whether the code and data are provided or not.
        \item If the contribution is a dataset and\slash or model, the authors should describe the steps taken to make their results reproducible or verifiable. 
        \item Depending on the contribution, reproducibility can be accomplished in various ways. For example, if the contribution is a novel architecture, describing the architecture fully might suffice, or if the contribution is a specific model and empirical evaluation, it may be necessary to either make it possible for others to replicate the model with the same dataset, or provide access to the model. In general. releasing code and data is often one good way to accomplish this, but reproducibility can also be provided via detailed instructions for how to replicate the results, access to a hosted model (e.g., in the case of a large language model), releasing of a model checkpoint, or other means that are appropriate to the research performed.
        \item While NeurIPS does not require releasing code, the conference does require all submissions to provide some reasonable avenue for reproducibility, which may depend on the nature of the contribution. For example
        \begin{enumerate}
            \item If the contribution is primarily a new algorithm, the paper should make it clear how to reproduce that algorithm.
            \item If the contribution is primarily a new model architecture, the paper should describe the architecture clearly and fully.
            \item If the contribution is a new model (e.g., a large language model), then there should either be a way to access this model for reproducing the results or a way to reproduce the model (e.g., with an open-source dataset or instructions for how to construct the dataset).
            \item We recognize that reproducibility may be tricky in some cases, in which case authors are welcome to describe the particular way they provide for reproducibility. In the case of closed-source models, it may be that access to the model is limited in some way (e.g., to registered users), but it should be possible for other researchers to have some path to reproducing or verifying the results.
        \end{enumerate}
    \end{itemize}

\item {\bf Open access to data and code}
    \item[] Question: Does the paper provide open access to the data and code, with sufficient instructions to faithfully reproduce the main experimental results, as described in supplemental material?
    \item[] Answer: \answerYes{} 
    \item[] Justification: The datasets used (Mozilla Common Voice, Multilingual LibriSpeech, ParlaSpeech) are publicly available, and the code with sufficient instructions to reproduce the main results is provided.
    \item[] Guidelines:
    \begin{itemize}
        \item The answer \answerNA{} means that paper does not include experiments requiring code.
        \item Please see the NeurIPS code and data submission guidelines (\url{https://neurips.cc/public/guides/CodeSubmissionPolicy}) for more details.
        \item While we encourage the release of code and data, we understand that this might not be possible, so \answerNo{} is an acceptable answer. Papers cannot be rejected simply for not including code, unless this is central to the contribution (e.g., for a new open-source benchmark).
        \item The instructions should contain the exact command and environment needed to run to reproduce the results. See the NeurIPS code and data submission guidelines (\url{https://neurips.cc/public/guides/CodeSubmissionPolicy}) for more details.
        \item The authors should provide instructions on data access and preparation, including how to access the raw data, preprocessed data, intermediate data, and generated data, etc.
        \item The authors should provide scripts to reproduce all experimental results for the new proposed method and baselines. If only a subset of experiments are reproducible, they should state which ones are omitted from the script and why.
        \item At submission time, to preserve anonymity, the authors should release anonymized versions (if applicable).
        \item Providing as much information as possible in supplemental material (appended to the paper) is recommended, but including URLs to data and code is permitted.
    \end{itemize}

\item {\bf Experimental setting/details}
    \item[] Question: Does the paper specify all the training and test details (e.g., data splits, hyperparameters, how they were chosen, type of optimizer) necessary to understand the results?
    \item[] Answer: \answerYes{} 
    \item[] Justification: All necessary experimental settings, including dataset splits, hyperparameters, optimizer choices, learning rates, scheduling, etc., are explicitly detailed in Section \ref{sec:experiments} and Appendices \ref{appendix_dataset_detail} and \ref{Appendix:Extra_Implemetation_Detail}.
    \item[] Guidelines:
    \begin{itemize}
        \item The answer \answerNA{} means that the paper does not include experiments.
        \item The experimental setting should be presented in the core of the paper to a level of detail that is necessary to appreciate the results and make sense of them.
        \item The full details can be provided either with the code, in appendix, or as supplemental material.
    \end{itemize}

\item {\bf Experiment statistical significance}
    \item[] Question: Does the paper report error bars suitably and correctly defined or other appropriate information about the statistical significance of the experiments?
    \item[] Answer: \answerNo{} 
    \item[] Justification: We omit error bars due to the prohibitive computational cost of repeated adversarial fine-tuning and evaluation across four large-scale ASR backbones. We instead evaluate the reported WER and CER metrics on sizeable test sets of $\approx$4 hours of speech per language.
    \item[] Guidelines:
    \begin{itemize}
        \item The answer \answerNA{} means that the paper does not include experiments.
        \item The authors should answer \answerYes{} if the results are accompanied by error bars, confidence intervals, or statistical significance tests, at least for the experiments that support the main claims of the paper.
        \item The factors of variability that the error bars are capturing should be clearly stated (for example, train/test split, initialization, random drawing of some parameter, or overall run with given experimental conditions).
        \item The method for calculating the error bars should be explained (closed form formula, call to a library function, bootstrap, etc.)
        \item The assumptions made should be given (e.g., Normally distributed errors).
        \item It should be clear whether the error bar is the standard deviation or the standard error of the mean.
        \item It is OK to report 1-sigma error bars, but one should state it. The authors should preferably report a 2-sigma error bar than state that they have a 96\% CI, if the hypothesis of Normality of errors is not verified.
        \item For asymmetric distributions, the authors should be careful not to show in tables or figures symmetric error bars that would yield results that are out of range (e.g., negative error rates).
        \item If error bars are reported in tables or plots, the authors should explain in the text how they were calculated and reference the corresponding figures or tables in the text.
    \end{itemize}

\item {\bf Experiments compute resources}
    \item[] Question: For each experiment, does the paper provide sufficient information on the computer resources (type of compute workers, memory, time of execution) needed to reproduce the experiments?
    \item[] Answer: \answerYes{} 
    \item[] Justification: We provide an ablation study comparing the training computational cost of our method and standard-AFT in Appendix~\ref{appendix:training_cost_comparison}.
    \item[] Guidelines:
    \begin{itemize}
        \item The answer \answerNA{} means that the paper does not include experiments.
        \item The paper should indicate the type of compute workers CPU or GPU, internal cluster, or cloud provider, including relevant memory and storage.
        \item The paper should provide the amount of compute required for each of the individual experimental runs as well as estimate the total compute. 
        \item The paper should disclose whether the full research project required more compute than the experiments reported in the paper (e.g., preliminary or failed experiments that didn't make it into the paper). 
    \end{itemize}
    
\item {\bf Code of ethics}
    \item[] Question: Does the research conducted in the paper conform, in every respect, with the NeurIPS Code of Ethics \url{https://neurips.cc/public/EthicsGuidelines}?
    \item[] Answer: \answerYes{} 
    \item[] Justification: Our research fully adheres to the NeurIPS Code of Ethics. We utilize publicly available datasets, and our work proactively aims to expose vulnerabilities and improve the security and trustworthiness of real-world ASR systems.
    \item[] Guidelines:
    \begin{itemize}
        \item The answer \answerNA{} means that the authors have not reviewed the NeurIPS Code of Ethics.
        \item If the authors answer \answerNo, they should explain the special circumstances that require a deviation from the Code of Ethics.
        \item The authors should make sure to preserve anonymity (e.g., if there is a special consideration due to laws or regulations in their jurisdiction).
    \end{itemize}

\item {\bf Broader impacts}
    \item[] Question: Does the paper discuss both potential positive societal impacts and negative societal impacts of the work performed?
    \item[] Answer: \answerYes{} 
    \item[] Justification: We discuss the societal impact in Section \ref{sec:introduction}, highlighting the risks of adversarial vulnerabilities in safety-critical ASR applications (e.g., in-vehicle voice interfaces) and how our proposed defense mitigates these threats.
    \item[] Guidelines:
    \begin{itemize}
        \item The answer \answerNA{} means that there is no societal impact of the work performed.
        \item If the authors answer \answerNA{} or \answerNo, they should explain why their work has no societal impact or why the paper does not address societal impact.
        \item Examples of negative societal impacts include potential malicious or unintended uses (e.g., disinformation, generating fake profiles, surveillance), fairness considerations (e.g., deployment of technologies that could make decisions that unfairly impact specific groups), privacy considerations, and security considerations.
        \item The conference expects that many papers will be foundational research and not tied to particular applications, let alone deployments. However, if there is a direct path to any negative applications, the authors should point it out. For example, it is legitimate to point out that an improvement in the quality of generative models could be used to generate Deepfakes for disinformation. On the other hand, it is not needed to point out that a generic algorithm for optimizing neural networks could enable people to train models that generate Deepfakes faster.
        \item The authors should consider possible harms that could arise when the technology is being used as intended and functioning correctly, harms that could arise when the technology is being used as intended but gives incorrect results, and harms following from (intentional or unintentional) misuse of the technology.
        \item If there are negative societal impacts, the authors could also discuss possible mitigation strategies (e.g., gated release of models, providing defenses in addition to attacks, mechanisms for monitoring misuse, mechanisms to monitor how a system learns from feedback over time, improving the efficiency and accessibility of ML).
    \end{itemize}
    
\item {\bf Safeguards}
    \item[] Question: Does the paper describe safeguards that have been put in place for responsible release of data or models that have a high risk for misuse (e.g., pre-trained language models, image generators, or scraped datasets)?
    \item[] Answer: \answerNA{} 
    \item[] Justification: Our work proposes a defensive framework to protect ASR models against adversarial attacks, which inherently acts as a safeguard. We do not release high-risk models or datasets intended for misuse.
    \item[] Guidelines:
    \begin{itemize}
        \item The answer \answerNA{} means that the paper poses no such risks.
        \item Released models that have a high risk for misuse or dual-use should be released with necessary safeguards to allow for controlled use of the model, for example by requiring that users adhere to usage guidelines or restrictions to access the model or implementing safety filters. 
        \item Datasets that have been scraped from the Internet could pose safety risks. The authors should describe how they avoided releasing unsafe images.
        \item We recognize that providing effective safeguards is challenging, and many papers do not require this, but we encourage authors to take this into account and make a best faith effort.
    \end{itemize}

\item {\bf Licenses for existing assets}
    \item[] Question: Are the creators or original owners of assets (e.g., code, data, models), used in the paper, properly credited and are the license and terms of use explicitly mentioned and properly respected?
    \item[] Answer: \answerYes{} 
    \item[] Justification: All prior works and existing assets (baseline models, datasets) are properly cited, and their open/permissive licenses (e.g., CC0, CC‑BY‑4.0, MIT) have been respected. Sources are given in the references.
    \item[] Guidelines:
    \begin{itemize}
        \item The answer \answerNA{} means that the paper does not use existing assets.
        \item The authors should cite the original paper that produced the code package or dataset.
        \item The authors should state which version of the asset is used and, if possible, include a URL.
        \item The name of the license (e.g., CC-BY 4.0) should be included for each asset.
        \item For scraped data from a particular source (e.g., website), the copyright and terms of service of that source should be provided.
        \item If assets are released, the license, copyright information, and terms of use in the package should be provided. For popular datasets, \url{paperswithcode.com/datasets} has curated licenses for some datasets. Their licensing guide can help determine the license of a dataset.
        \item For existing datasets that are re-packaged, both the original license and the license of the derived asset (if it has changed) should be provided.
        \item If this information is not available online, the authors are encouraged to reach out to the asset's creators.
    \end{itemize}

\item {\bf New assets}
    \item[] Question: Are new assets introduced in the paper well documented and is the documentation provided alongside the assets?
    \item[] Answer: \answerNA{} 
    \item[] Justification: Our paper does not introduce new datasets or assets; we utilize publicly available datasets and open-source models.
    \item[] Guidelines:
    \begin{itemize}
        \item The answer \answerNA{} means that the paper does not release new assets.
        \item Researchers should communicate the details of the dataset\slash code\slash model as part of their submissions via structured templates. This includes details about training, license, limitations, etc. 
        \item The paper should discuss whether and how consent was obtained from people whose asset is used.
        \item At submission time, remember to anonymize your assets (if applicable). You can either create an anonymized URL or include an anonymized zip file.
    \end{itemize}

\item {\bf Crowdsourcing and research with human subjects}
    \item[] Question: For crowdsourcing experiments and research with human subjects, does the paper include the full text of instructions given to participants and screenshots, if applicable, as well as details about compensation (if any)? 
    \item[] Answer: \answerNA{} 
    \item[] Justification: Our research does not involve any crowdsourcing or human subjects.
    \item[] Guidelines:
    \begin{itemize}
        \item The answer \answerNA{} means that the paper does not involve crowdsourcing nor research with human subjects.
        \item Including this information in the supplemental material is fine, but if the main contribution of the paper involves human subjects, then as much detail as possible should be included in the main paper. 
        \item According to the NeurIPS Code of Ethics, workers involved in data collection, curation, or other labor should be paid at least the minimum wage in the country of the data collector. 
    \end{itemize}

\item {\bf Institutional review board (IRB) approvals or equivalent for research with human subjects}
    \item[] Question: Does the paper describe potential risks incurred by study participants, whether such risks were disclosed to the subjects, and whether Institutional Review Board (IRB) approvals (or an equivalent approval/review based on the requirements of your country or institution) were obtained?
    \item[] Answer: \answerNA{} 
    \item[] Justification: Our paper does not involve human subjects, making IRB approvals inapplicable to our work.
    \item[] Guidelines:
    \begin{itemize}
        \item The answer \answerNA{} means that the paper does not involve crowdsourcing nor research with human subjects.
        \item Depending on the country in which research is conducted, IRB approval (or equivalent) may be required for any human subjects research. If you obtained IRB approval, you should clearly state this in the paper. 
        \item We recognize that the procedures for this may vary significantly between institutions and locations, and we expect authors to adhere to the NeurIPS Code of Ethics and the guidelines for their institution. 
        \item For initial submissions, do not include any information that would break anonymity (if applicable), such as the institution conducting the review.
    \end{itemize}

\item {\bf Declaration of LLM usage}
    \item[] Question: Does the paper describe the usage of LLMs if it is an important, original, or non-standard component of the core methods in this research? Note that if the LLM is used only for writing, editing, or formatting purposes and does \emph{not} impact the core methodology, scientific rigor, or originality of the research, declaration is not required.
    \item[] Answer: \answerNA{} 
    \item[] Justification: The core methodological development in this research does not involve the use of LLMs as important, original, or non-standard components. LLMs were not used in the design or evaluation of the proposed methods.
    \item[] Guidelines:
    \begin{itemize}
        \item The answer \answerNA{} means that the core method development in this research does not involve LLMs as any important, original, or non-standard components.
        \item Please refer to our LLM policy in the NeurIPS handbook for what should or should not be described.
    \end{itemize}

\end{enumerate}

\end{document}